\documentclass[pdflatex,sn-mathphys-num]{sn-jnl}

\usepackage{graphicx}%
\usepackage[strings]{underscore}
\usepackage{multirow}%
\usepackage{amsmath,amssymb,amsfonts}%
\usepackage{amsthm}%
\usepackage{mathrsfs}%
\usepackage[title]{appendix}%
\usepackage{xcolor}%
\usepackage{textcomp}%
\usepackage{manyfoot}%
\usepackage{booktabs}%
\usepackage{algorithm}%
\usepackage{algorithmicx}%
\usepackage{algpseudocode}%
\usepackage{listings}%
\usepackage{lineno}%

\theoremstyle{thmstyleone}%
\theoremstyle{thmstyletwo}%

\theoremstyle{thmstylethree}%

\begin{document}

\title[Article Title]{Deep Learning for predicting rate-induced tipping}


\author*[1,2]{\fnm{Yu} \sur{Huang}}\email{y.huang@tum.de}

\author[1,2]{\fnm{Sebastian} \sur{Bathiany}}

\author[3]{\fnm{Peter} \sur{Ashwin}}

\author[1,2,3]{\fnm{Niklas} \sur{Boers}}

\affil[1]{\orgdiv{Earth System Modelling}, \orgname{School of Engineering and Design, Technical University of Munich}, \city{Munich}, \postcode{80333}, \country{Germany}}

\affil[2]{\orgdiv{Complexity Science}, \orgname{Potsdam Institute for Climate Impact Research}, \city{Potsdam}, \postcode{14473}, \country{Germany}}

\affil[3]{\orgdiv{Department of Mathematics and Statistics}, \orgname{University of Exeter}, \city{Exeter}, \postcode{EX4 4QF}, \country{United Kingdom}}


\abstract{Nonlinear dynamical systems exposed to changing forcing can exhibit catastrophic transitions between alternative and often markedly different states. The phenomenon of critical slowing down (CSD) can be used to anticipate such transitions if caused by a bifurcation and if the change in forcing is slow compared to the internal time scale of the system. However, in many real-world situations, these assumptions are not met and transitions can be triggered because the forcing exceeds a critical rate.
For example, given the pace of anthropogenic climate change in comparison to the internal time scales of key Earth system components, such as the polar ice sheets or the Atlantic Meridional Overturning Circulation, such rate-induced tipping poses a severe risk. Moreover, depending on the realisation of random perturbations, some trajectories may transition across an unstable boundary, while others do not, even under the same forcing. CSD-based indicators generally cannot distinguish these cases of noise-induced tipping versus no tipping. This severely limits our ability to assess the risks of tipping, and to predict individual trajectories. To address this, we make a first attempt to develop a deep learning framework to predict transition probabilities of dynamical systems ahead of rate-induced transitions.
Our method issues early warnings, as demonstrated on three prototypical systems for rate-induced tipping, subjected to time-varying equilibrium drift and noise perturbations. Exploiting explainable artificial intelligence methods, our framework captures the fingerprints necessary for early detection of rate-induced tipping, even in cases of long lead times. Our findings demonstrate the predictability of rate-induced and noise-induced tipping, advancing our ability to determine safe operating spaces for a broader class of dynamical systems than possible so far.}




\maketitle

\section{Introduction}

Tipping points denote critical thresholds in nonlinear dynamical systems, where changing environmental conditions can lead to a collapse into distinctly different states as a bifurcation point is crossed. This phenomenon has been substantiated through theoretical and observational studies in diverse real-world systems, encompassing the climate system \cite{dakos2008slowing, ashwin2012tipping, boers2022theoretical}, ecosystems \cite{scheffer2001catastrophic, barnosky2012approaching, flores2024critical}, financial crises \cite{faranda2015early}, or the human brain \cite{maturana2020critical}. Therefore, a central objective is to detect early warning signals (EWS) that the system may be approaching a tipping point \cite{scheffer2009early, lenton2011early, lohmann2021risk, armstrong2022exceeding}. 

Dynamical systems theory suggests that when slowly varying external forcing is far from a bifurcation point \cite{scheffer2009early, ashwin2012tipping,boers2022theoretical}, often indicated by a threshold value of the forcing, the state of the system remains in the basin of attraction of the quasi-static state, and after any minor perturbation the system will promptly return to its equilibrium state. As a bifurcation-induced tipping is approached, the basin of attraction undergoes a reduction in its curvature (local stability). As a consequence, even slight perturbations begin to exhibit a more prolonged effect in the dynamics, referred to as critical slowing down (CSD) \cite{dakos2008slowing, scheffer2009early, boers2021observation}. Particularly in the presence of noise perturbations, this manifests as an increase in variance and lag-1 autocorrelation in time series data, serving as statistical EWS that often precedes bifurcation-induced tipping events across various systems. Recent advancements have showcased the efficacy of deep learning (DL) techniques in offering EWS for bifurcation-induced tipping \cite{bury2021deep, patel2023using}. 

However, in many cases the changing rate of external forcing is too rapid for the system to maintain its quasi-equilibrium state. In such situations the collapse can occur unexpectedly, even if there is no bifurcation tipping \cite{ashwin2012tipping, lohmann2021risk}. This rate-induced tipping (R-tipping) has been documented across a wide range of dynamical systems \cite{lohmann2021risk, panahi2023rate, ritchie2023rate}. In particular in the context of anthropogenic climate change, it is likely that the forcing rate is fast compared to the characteristic time scales of key Earth system components that have been suggested to exhibit tipping potential \cite{lenton2019climate, lohmann2021risk, boers2022theoretical, armstrong2022exceeding}. Unlike bifurcation-induced tipping, R-tipping is not linked to a loss of stability of equilibrium, and the shape of the system’s basin of attraction can remain invariant. Instead, it shifts at an accelerating rate (Fig. 1). In particular, amid minor disturbances such as noise or within intricate high-dimensional systems, anticipating the occurrence of R-tipping poses a challenge. If the rate of the forcing is close to the critical rate required for R-tipping, it depends on the noise realisation whether an individual trajectory stays within the basin of attraction of the autonomous system, or undergoes a transition across its boundary. Such transitions into an alternative basin can even occur without any change in forcing and have been named noise-induced tipping (N-tipping), see \cite{ashwin2012tipping}. Therefore, the challenge in predicting R-tipping arises from the simultaneous involvement of rapidly changing forcing signals and N-tipping mechanisms \cite{lohmann2021risk, ritchie2016early, ritchie2017probability, slyman2023rate}.

As depicted in Fig. 2A, during experimentation on a prototypical R-tipping system, using ensemble simulations with identical time-varying forcing but different instant noise perturbations, part of the realizations exhibit tipping, while others do not. The randomness of perturbations precludes the inference of whether a specific realization will undergo tipping based on the system's equations alone. In this case, the occurrence of R-tipping also lacks deterministic dependence on the changing rate of forcing, as evidenced by the broad distribution of R-tipping occurrence times (see Materials and Methods) illustrated in the bottom panel of Fig. 2A. Moreover, based on visual observations, the time series realizations that exhibit R-tipping under noise influence cannot be discriminated from those that do not exhibit tipping prior to the transition, and it is unclear if CSD can distinguish between R-tipping with non-tipping realizations. So far, identifying EWS for R-tipping under noise perturbations still proves exceptionally challenging. It is noted that when systems approach tipping points, the phenomena and dynamical mechanisms could become common to both low-dimensional and high-dimensional systems \cite{kuznetsov1998elements, campbell2009calculating, lohmann2021risk}. For instance, the abovementioned issues about tipping phenomena unexpected by bifurcation theory were also underscored in ensemble simulation experiments pertaining to the Atlantic Meridional Overturning Circulation (AMOC) in climate studies \cite{lohmann2021risk, romanou2023stochastic, cini2024simulating}. Consequently, these observations raise considerable concern about the ambiguous safe operating space under the risks of R-tipping, especially in policies addressing anthropogenic climate change \cite{lohmann2021risk, ritchie2021overshooting, wunderling2023global}, and enhancing the resilience of ecosystems \cite{smith2022empirical, panahi2023rate, flores2024critical}. In the context of anthropogenic climate change, the risk of R-tipping may have been greatly underestimated, and to date no method to anticipate such transitions had been proposed.

Here we aim to improve this situation by predicting R-tipping occurrence among an ensemble of noise perturbations, using deep learning. We first present a composite analysis to demonstrate that CSD cannot serve as the indicator for discerning R-tipping amidst time-varying external forcing and noise perturbations. Thereafter, we show that the application of an interpretable deep learning framework can discern the subtle hidden information beyond CSD purely from the data, which enables the detection of the early warning fingerprints preceding R-tipping.

\section{Results}

\subsection*{Potential predictability of R-tipping} 
To reproduce the dynamics of R-tipping, we implement three prototypical example systems from different scientific disciplines. First, we examine a prototype system for R-tipping introduced by Ashwin et al. \cite{ashwin2012tipping} (Saddle–node system), incorporating additive noise perturbations \cite{ritchie2016early} (see Materials and Methods). When the system’s basin of attraction shifts slowly due to a time-varying environmental forcing, the system can follow the shifting attraction basin and promptly recover toward the equilibrium. Conversely, a very rapid shift of the attractor can cause the system to escape from the basin, leading to R-tipping (Fig. 1B). There exist a theoretical threshold value $\epsilon_c$ for the shift rate that triggers R-tipping \cite{ashwin2012tipping, ritchie2016early}. In cases where a shifting basin of attraction and noise perturbations coexist, tipping can manifest even when the shift rate is not as fast as $\epsilon_c$ (Fig. 1C). The results presented in Fig. 2A align with such a scenario, and the statistics are derived from 300,000 ensemble realizations, with approximately 37\% of them experiencing R-tipping. Specifically, we select 60,000 time series with R-tipping (group A) and an additional 60,000 without tipping (group B) for further analysis. 

Based on the time of R-tipping occurrences (as defined in the Materials and Methods), for group A, we retain the time series segments prior to this tipping time. To ensure fairness in statistics, the cut-off point for each time series segment in group B aligns with the corresponding point in group A. Subsequently, we calculate the classical CSD indicators autocorrelation and variance (see Materials and Methods) for each time series and record the composite mean values and 99th percentiles of these metrics within groups A and B respectively. Results show that the autocorrelations within groups A and B cannot be distinguished over time, evident in both their composite mean values and 99\% confidence intervals (Fig. 3A). Regarding the variance, the composite mean values for groups A and B both display increasing trends over time, and the 99\% confidence intervals for them substantially overlap. 
Hence, unlike for the case of bifurcation-induced tipping, CSD cannot discern R-tipping from the non-tipping cases amidst the changing forcing and noise perturbations. CSD can, hence, not serve to anticipate R-tipping. This may be expected, since CSD focuses on changes in the linear restoring rate and how that affects the autocorrelation or variance; the underlying assumption that the system remains close to equilibrium and that the linearization around the equilibrium is valid, is by design broken in the R-tipping context.

We further employ two additional prototype R-tipping systems with different internal dynamics to conduct similar composite analyses for two additional prototypical models, namely the normal form of the Bautin system \cite{alkhayuon2018rate} and the Compost-bomb system \cite{luke2011soil} (see details in the following sections), revealing that CSD continues to fail to anticipate the transitions (Figs. 3B and 3C). Note that the increase in autocorrelation and variance may indicate that these two measures reflect the systems' deviation from equilibrium; however, the way they increase for both tipping and non-tipping cases implies that this information cannot be used to predict if a transition occurs. Moreover, the CSD theory relies on linearizing the dynamics around a given stable equilibrium, so using CSD indicators in cases where the system is not close to equilibrium is not mathematically justified. 
 
 We thus search for alternative ways to identify a precursor signal in time series prior to the R-tipping occurrence. To this end, we first examine the probability distribution of time series values over 100 time steps preceding the onset of R-tipping for the Saddle–node system (SI Appendix, Fig. S1A), revealing significant differences between R-tipping and non-tipping scenarios preceding the R-tipping occurrence. While there is overlap in the probability distributions for these two scenarios, distinctions can be observed in the profiles of the probability distributions. We then employ a Kolmogorov-Smirnov significance test to examine whether the probability distribution under the R-tipping scenario is distinguishable from that under the non-tipping scenario. This analysis is performed across various lead times preceding the occurrence of R-tipping (SI Appendix, Fig. S2A). It is found that significant differences in probability distributions between R-tipping and non-tipping scenarios persist up to 280 time steps before the onset of R-tipping. Similar conclusions are also evident in the analyses conducted for the Bautin system and the Compost-bomb system (SI Appendix, Figs. S1 and S2). This suggests the presence of higher-order statistical information beyond CSD, which can differentiate between R-tipping and non-tipping time series, thus making R-tipping potentially predictable. We conjecture that these characteristic higher-order statistics represent how far the system in question is away from equilibrium. This encourages us to pursue further investigations to identify valid precursor signals for R-tipping.

\subsection*{Predicting R-tipping probability by DL}
Distinguishable features are evident in the probability distributions of ensemble time series prior to the R-tipping occurrence. However, this information alone does not provide a direct way for determining whether a single time series will exhibit R-tipping under additional noise perturbations. Hence, we hypothesize that feeding these ensemble time series into deep learning (DL) models will enable extraction of essential features from both the probability distributions and time series structures. This, in turn, would facilitate inference on whether a single time series will manifest R-tipping or not. Aiming to establish a DL-based indicator for predicting R-tipping, our DL models integrate a Convolutional Neural Network (CNN) with a fully-connected Neural Network, enabling the extraction of both local and global information from time series \cite{ismail2019deep}. For a specific lead time before the R-tipping occurrence, we correspondingly train a DL model to discern between R-tipping and non-tipping scenarios, such that we can specifically inspect the potential difference between R-tipping and non-tipping scenarios at each forecast lead time. When feeding a time series segment into the trained DL model, the binary outputs represent the probabilities of this time series, at this lead time, to be an R-tipping scenario or a non-tipping scenario, respectively (SI Appendix, Fig. S3). Detailed DL model configurations and training settings are described in the Materials and Methods section. Here, the probability output by the trained DL models is taken to provide a prediction indicator, explaining the real-time probability of the system state to approach R-tipping. 

As seen in SI Appendix, Fig. S4A, comparing two given time series at a given lead time before the R-tipping occurs, visual or CSD-based inspection if they will experience R-tipping is not feasible. CSD indicators show increasing trends in autocorrelation and variance for both R-tipping and non-tipping cases. In contrast, our DL model predicts an increase in the R-tipping probability for the time series under R-tipping scenario, but a decrease for the time series in the non-tipping scenario. Iterating predictions using the trained DL models on time series within the aforesaid groups A and B, the composite results show that this DL-derived R-tipping indicator can clearly distinguish R-tipping (group A) from non-tipping (group B) scenarios, with a long lead time before the R-tipping (bottom panel of Fig. 3A). At each forecast lead time, we employ the Kolmogorov-Smirnov significance test to assess whether the R-tipping probability of Group A are distinguishable from those of Group B (see SI Appendix, bottom panel of Fig. S2A). The findings reveal that up to 290 time steps before the onset of R-tipping in the Saddle–node system, the DL-derived R-tipping probabilities in the two groups have become distinguishable. This observation aligns with the analysis results regarding the probability distributions of system states (SI Appendix, top panel of Fig. S2A). Consistent conclusions can be drawn regarding the Bautin system and Compost-bomb system (refer to SI Appendix, bottom panels of Figs. S2B and S2C). According to Kolmogorov-Smirnov test results, their R-tipping probabilities can be discerned, respectively, at 130 and 1000 time steps prior to the onset of R-tipping.       

The DL prediction results are further assessed via two additional aspects. Firstly, as an alternative to forecasting, one can randomly choose a time series from either group A or group B and predict whether it will undergo R-tipping. In this case, the anticipated probability of the chosen time series experiencing R-tipping is approximately 50\% by construction of the time series ensembles, aligning with an expected accuracy rate of 50\% for the prediction. Thus, if the R-tipping probability predicted by the DL model is significantly different from 50\%, it indicates that the prediction is valid. For both groups A and B, starting from around 200 time steps before the R-tipping, their respective ensemble distributions start to spread and diverge from the baseline of 50\%. This observation indicates that the DL model demonstrates predictive skills from that time onward, offering an informative long-lead forecast. 

The second straightforward prediction strategy involves examining the composited time series envelopes of groups A and B (refer to the top panel of Fig. 3A): it becomes apparent that the 99\% confidence intervals of R-tipping and non-tipping scenarios begin to partially separate before the actual onset of R-tipping. Thus if a particular time series begins to surpass the 99\% confidence interval of the non-tipping scenario, it could serve as an indicative signal of an impending R-tipping. Accordingly, we can estimate the accuracy of this prediction strategy by counting the number of group-A cases that fall outside the envelope of group B at each forecast lead time (SI Appendix, Fig. S5A). Specifically, between 100 and 10 time steps before R-tipping of the Saddle–node system, the success rate of envelope-based prediction consistently remains below 20\%, while concurrently, the success rate of the trained DL models steadily rises from 60\% to 95\% (SI Appendix, Fig. S5B). Similarly, when forecasting R-tipping in the other two systems, our DL approach clearly outperforms the envelope-based method (refer to SI Appendix, Figs. S5D and S5F).

It is noteworthy that the predictive accuracy of the DL models shows weak disparity between the training and testing phases (SI Appendix, Fig. S5), indicating that overfitting is not an issue for our DL model. This observation suggests that the trained DL model has robustly learned the predictive information of R-tipping. The predictive capability of our DL model hence markedly surpasses that of CSD and the ensemble time series envelopes, demonstrating that the model has successfully extracted higher-order statistical information from the time series data. 

The feasible predictability of R-tipping can be further substantiated in two additional prototype systems. The Bautin system features R-tipping in a dynamically different way than the Saddle-node system, simulating a periodic attractor that undergoes shifts due to changing external forcing \cite{alkhayuon2018rate}. In addition, the Compost-bomb system embodies a prototypical interaction between soil carbon and climate change, capable of exhibiting R-tipping in the context of dynamical feedbacks \cite{luke2011soil} (see Materials and Methods). With these two systems we can investigate R-tipping predictability under conditions with much more internal variability (see the example time series in SI Appendix, Fig. S4). Despite different internal dynamics compared to the Saddle–node system, the combined impact of rapid external forcing and noise perturbations triggering R-tipping, is common to all three systems. Similar to the observations in the Saddle–node system, the occurrences of R-tipping in the Bautin system and the Compost-bomb system are subtle and uncertain under their respective dynamics, particularly in the presence of noise perturbations (Fig. 2). Also for these latter two systems, classical CSD indicators fail to anticipate R-tipping, as expected. Employing a consistent DL configuration and training strategy, the trained DL models, in contrast, provide reliable long-lead forecasts for R-tipping within the Bautin system and the Compost-bomb system (Fig. 3 and SI Appendix, Fig. S3). This demonstrates that predicting R-tipping trajectories is possible in diverse dynamical systems.

\subsection*{Fingerprints and variability of R-tipping probabilities}
Our DL model exhibits a notable ability to differentiate between R-tipping and non-tipping scenarios in advance of the R-tipping occurrence, enabling to predict such R-tipping cases. This sparks further interest to understand the DL model predictions, e.g. regarding the question if there is a particular time window during which the system's state significantly influences the probability of subsequent R-tipping. This hypothesis can be tested by monitoring the pivotal features within the input time series that contribute to the DL model's ultimate predictions. Here this analysis is facilitated by the Layer-wise Relevance Propagation (LRP) algorithm, guiding interpretation of DL models \cite{bach2015pixel, montavon2019layer} (see Materials and Methods). For each forecast lead time, the LRP scores can be considered as a function of time, indicating the relative importance of each time point within the input time series for guiding the prediction of the DL model. We compute the LRP scores for the trained DL models as they process time series from groups A and B, respectively. Upon comparing the 99\% confidence intervals of the two groups of LRP scores, evident distinctions emerge in the attention patterns of DL models between R-tipping and non-tipping scenarios (blue and grey shading areas in Fig. 4). Regardless of the chosen forecast lead time, the curves of LRP scores consistently exhibit maximum peaks near the end of the time series. This suggests that the R-tipping probability of the system primarily relies on its past neighboring dynamic states. The position of this R-tipping fingerprint varies with lead time and depends on the system's instantaneous dynamical evolution. 

In the case of the Saddle-node system, another R-tipping fingerprint can be observed: when the forecast lead time is set to 0, 30, and 50 respectively, the curves of LRP scores all display peaks within the time window ranging from -100 to -50 time steps (refer to Figs. 4C, 4D, and 4E), which are absent in the LRP score profiles of the non-tipping scenario. An examination of the LRP scores of a sample time series (example 1 in Fig. 4) reveals the close relationship between its temporal R-tipping probability evolution and the pattern of LRP scores: When its LRP score profiles closely resemble those of the R-tipping scenario (Figs. 4C and 4D), the DL model infers that its R-tipping probability exceeds 50\% (see Fig. 4B), and vice versa (see Fig. 4E). However, when the forecast lead time is extended to 150 time steps, the differences in the LRP scores' patterns between R-tipping and non-tipping scenarios become smaller (Fig. 4F). At this forecast lead time, the R-tipping probability is approximately 50\% (Fig. 4B), indicating weaker predictability of R-tipping. Similarly, in the Bautin system and Compost-bomb system, the LRP analyses also demonstrate informative fingerprints of R-tipping prior to its onset (refer to SI Appendix, Figs. S6 and S7), and the changes in LRP scores are consistent with the R-tipping probability.

Examining the variability of the estimated R-tipping probabilities aids in a further understanding of the DL model predictions. We further compare different time series in the R-tipping scenario and find that if the trajectory of a time series approaches the upper boundary of the R-tipping ensemble envelope (example 1 in Fig.4A), its R-tipping probability reaches as high as 70\% within the time range of -100 and -50 time steps before tipping (Fig.4B). For the time series approaching the lower boundary (example 3 in Fig.4A), its R-tipping probability stays lower than 50\% over time, and starts to increase only during the last 50 time steps prior to tipping. For time series situated within the envelope center (example 2 in Fig.4A), its R-tipping probability curve over time lies between those of the former two cases. These statistical measures exhibit sensitive changes around the consistent periods with the aforesaid fingerprint. Considering the dynamical equation of the Saddle–node system, this variability in R-tipping probability is attributed to two factors: forcing and noise perturbations. We investigate their effects on the R-tipping probability. As shown in Fig. 2A, the instant shift rate of the forcing (i.e., slope of the forcing) reaches its maximum at time = 0, yet R-tipping typically occurs with a delay relative to time step 0, and different R-tipping realizations have different delay time \cite{ritchie2016early}. To illustrate the impact of this delay effect on the R-tipping probability, the time series samples in group A are categorized into three equal quantiles based on the delay of their tipping time relative to the maximum of the instant shift rate. The DL-derived R-tipping probabilities within the quantiles are analyzed separately (SI Appendix, Fig. S8). When the time delay between R-tipping and the maximum of the shift rate is small (indicating relatively early tipping in dynamics), the estimated R-tipping probabilities consistently remain above 45\% over time, with much narrower confidence intervals. Conversely, with a larger time delay (suggesting relatively late tipping), the R-tipping probabilities may initially decrease to 25\% between time steps -100 and -50, subsequently rise to 100\% before R-tipping, and the associated confidence intervals are much broader compared to the case of early tipping. Hence, the specific dynamical process, whether it involves early or late tipping relative to the time of maximum shift rate, can influence the R-tipping probabilities in a long-lead forecast. Additionally, we note that the maximum of the shift rate falls exactly within the intervals of -100 to -50 time steps (SI Appendix, Fig. S8), coinciding with the position of the aforementioned fingerprints. This indicates that the DL model is sensitive to the intrinsic dynamical characteristics of the system embedded within the data.

The magnitude of noise perturbations is another factor contributing to the variability of the R-tipping probabilities. Using the Saddle–node system as an illustrative example, when experimenting with higher magnitudes of noise perturbations integrated into the system, the mean values of the estimated R-tipping probabilities overall decrease, and the distribution ranges narrow (SI Appendix, Fig. S9A). Consistent evidence from experiments on the Bautin system and the Compost-bomb system also supports these findings (SI Appendix, Fig. S9). This implies that as noise perturbations increase, the achievable predictability of R-tipping by our DL models diminishes.

\subsection*{Prediction accuracy across out-of-sample forcing rates}
Next, we asses the performance of our DL models in out-of-sample predictions, i.e. to predict R-tipping for forcing rates not encountered during training. While the above DL models are trained exclusively on time series with a specific forcing rate ($\epsilon=1.25$), we employ the trained models to predict R-tipping cases with different forcing rates, ranging from $\epsilon=0.9$ to $\epsilon=1.9$ (see the forcing time series in SI Appendix, Fig. S10). The prediction accuracy as a function of the forcing rate and forecast lead time for the Saddle-node system (Fig. 5A) demonstrates that the trained DL models can well adapt to the out-of-sample forcing scenarios. With a forecast lead time of 50 time steps, the DL model exhibits higher accuracy in cases featuring lower forcing rates. This can be explained by the fact that time series with lower forcing rates experience earlier increases compared to those with higher forcing rates (refer to SI Appendix, Fig. S10A). In support of this suggestion, we replace a new training dataset with a higher forcing rate $\epsilon=1.7$, and repeat the out-of-sample experiment. As in Figs. 5A and 5D, the prediction accuracy under these two training settings correspondingly shows consistent patterns. Similar conclusions can be drawn from experiments using the Compost-bomb system (Figs. 5C and 5F). However, in the case of the Bautin system, out-of-sample predictions are not universally successful across all unseen forcing scenarios. The DL models demonstrate high prediction accuracy primarily for forcing rates that are close to the training forcing rate (Figs. 5B and 5E). This could be attributed to the Bautin system displaying more complex dynamical responses to various forcing rates, as discussed earlier \cite{alkhayuon2018rate}, and highlights that the generaizability of our methods depends on the system under study.

\section{Discussion}

In contrast to the extensive literature on bifurcation-induced tipping, there remains a notable scarcity of methods for anticipating rate-induced tipping. In particular, the well-established CSD framework cannot be used when the rate of forcing is fast compared to the characteristic time scale of the system in question, since it assumes that the system dynamics can be linearized around that equilibrium. To fill this gap we introduced here a skillful deep learning based indicator to predict R-tipping amid noise perturbations. To test its performance we used datasets from prototypical R-tipping systems, affirming the predictability of R-tipping in the presence of noise perturbations with our method. The findings suggest that the DL algorithm can extract high-order statistical information quantifying how far a system is away from equilibrium and hence, how close it is to crossing the boundary of a given basin of attraction. This information can be readily used to give quantitative probabilities that a R-tipping event is forthcoming, and hence for predicting such an event. 

It is worth noting that besides R-tipping, there are other tipping phenomena such as global bifurcations \cite{ashwin2012tipping, adamson2020forecasting} and tipping transitions from non-equilibrium attractors \cite{xu2023non} that do not rely on changes to the local stability of equilibrium, and would hence also not be captured by CSD. We prospect that a comprehensive DL model could be taken to provide precursor signals across diverse tipping phenomena, and also distinguish between them in terms of their different forcing scenarios, purely based on the data. Achieving this would involve training the model on an extended dataset that encompasses those dynamics. By doing so, such a general DL model for tipping prediction should be expected to better identify the boundary of the safe operating space for dynamical systems, in terms of the position of critical forcing thresholds as well as critical rates of forcing. This would enable a comprehensive quantification of tipping risk in natural systems.

By utilizing interpretable DL techniques, we identified the presence and location of the precursory fingerprint of R-tipping in a time series; however, we did not delve deeply into the characteristics of this fingerprint. This limitation is partly attributed to the one-dimensional nature of the time series data, which restricts visual observations. A more in-depth analysis may necessitate the application of high-order statistical theories and techniques, for instance \cite{ashkenazy2003nonlinearity, bury2020detecting, he2021decreasing}. It can be speculated that for a three-dimensional field, such as in studies on the AMOC \cite{jackson2020fingerprints}, the characteristics of an R-tipping fingerprint could be presented in both spatial and temporal dimensions, allowing for a more nuanced observation and understanding of its nature and content. The adequate disclosure of precursory fingerprints for tipping points is valuable for further exploring and understanding the subtle dynamics of complex systems \cite{feng2014deep, jackson2020fingerprints, boers2021observation, ben2023uncertainties}. Leveraging interpretable deep learning techniques would greatly benefit these efforts.

For the time series data considered here, our model utilizes a framework based on one-dimensional convolutional neural networks and employs the LRP approach for interpreting the results. We also also experimented with using Multi-Layer Perceptron \cite{ismail2019deep} and Transformer \cite{wen2022transformers} architectures for the DL model. While they also achieved satisfactory predictive performance for R-tipping (not shown here), a comparative analysis suggested that our current framework performs better. Nevertheless, this suggest that our results are robust and not specific to the exact model architecture. When dealing with high-dimensional data, such as in the study of three-dimensional spatiotemporal fields like AMOC, more complex frameworks such as Transformers \cite{zhou2023self} could be considered. 

An important consideration in our study is that we specifically observe the predictability and fingerprint of R-tipping amidst noise perturbations at different time points before its occurrence. This theoretically allows for a better exploration of the predictability of R-tipping. However, this assumes that we already know the time of R-tipping occurrence. In practical predictions, we may not be aware of how far our prediction time point is from the occurrence of R-tipping. To address this issue in future work, one might focus in more detail on the results of the LRP application, which carry additional information. This approach would require more specific knowledge and description of the characteristics of the fingerprints. Moreover, the DL model framework could be modified by allowing the states of neurons in the model to update continuously with new time points of the input time series, thus to provide continuous predictions of R-tipping probabilities for various time points in the time series, which would place higher demands on the information extraction capabilities of the model framework. 


In summary, many real-world systems, such as ecosystems or climate subsystems, may be prone to tipping induced by a combination of rate- and noise-induced effects. For example, given the pace of anthropogenic climate change, it may be argued that combinations of rate- and noise-induced effects are more relevant than bifurcation-induced effects when assessing tipping risk in the climate change context. Noise perturbations may trigger transitions even in cases where the forcing of a system under study is below the critical rate. To our knowledge, no method for predicting transitions in such situations exists to date. We have introduced here a deep learning based method for this purpose, and have shown in three paradigmatic examples that our method is skillful in predicting such rate-induced tipping under the influence of noise perturbations. 

\section{Materials and Methods}

\subsection*{Prototype systems for R-tipping}

We examined the R-tipping dynamics using data from three paradigmatic systems, each manifesting R-tipping under different internal dynamical variability. The equations governing these systems are given here, which have been integrated to provide time series data for our comprehensive analyses. The first system stems from the normal form for Saddle-node bifurcation \cite{ashwin2012tipping, ritchie2016early}, incorporating additional noise perturbations: 
\begin{align*}
dX_t=[(X_t+\lambda(t))^2-1]dt+\sqrt{2D_{1}}dW_{1 t}.
\end{align*}
The system state $X_t$ is shifted by the forcing term $\lambda(t)$, and it follows $\lambda(t)=0.5\lambda_{max}[\tanh(0.5\lambda_{max}\epsilon t)+1]$, in which $\lambda_{max}=3$ and the time-varying rate of $\lambda(t)$ is controlled by $\epsilon$ (i.e., forcing rate). The white noise term $dW_{1 t}$, increments of a Wiener process $W_{1}$, represents the slight perturbation, and $D_{1}=0.008$ determines its magnitude. In absence of noise perturbations, $\epsilon_c=4/3$ determines the threshold of $\epsilon$ for triggering R-tipping. But in presence of noise perturbations, R-tipping can happen by chance when $\epsilon$ takes a smaller value than $\epsilon_c$, and here $\epsilon$ is set to 1.25 in order to allow such phenomena. 

The employed Bautin system involves a branch of periodic attractors \cite{alkhayuon2018rate}, subjected to additional noise perturbations, as follows:
\begin{align*}
dZ_t=&(a+i\omega)[Z_t-\Lambda(t)]dt-b[Z_t-\Lambda(t)]^2 [Z_t-\Lambda(t)]dt\\
&+[Z_t-\Lambda(t)]^4 [Z_t-\Lambda(t)]dt+D_{2}dW_{2 t},
\end{align*}
where the complex number $Z_t=X_t+Y_ti$ denotes the system state, and $\Lambda(t)$ the forcing term, following $\Lambda(t)=0.5\Lambda_{max}[\tanh(0.5\Lambda_{max}rt)+1]$. The forcing rate is controlled by $r$, with $r$ set to a small value of 0.1, far from the threshold required to trigger R-tipping in the absence of noise. $D_{2}=0.2$ determines the magnitude of white noise $dW_{2 t}$ that slightly perturbs the system. The rest of parameters is set to $\Lambda_{max}=8$, $a=0.1$, $\omega=3$ and $b=1$. 

The third system is the Compost-bomb model \cite{luke2011soil}, also with additional noise perturbations applied to the system state:
\begin{align*}
&dC_t=\Pi{}dt-C_tr_0e^{\alpha{}t}dt\\
&\mu{}dT_{t}=C_{t}r_{0}e^{\alpha{}t}Adt-\lambda{}[T_t-T_a(t)]dt+D_{3}dW_{3 t}
\end{align*}
For this system, $T_a(t)$ denotes the forcing from atmospheric temperature warming, following $T_a(t)=vt$. $T_t$ is the soil temperature, which undergoes a coupled feedback processes with soil respiration rate $r_0e^{\alpha{}t}$ and soil carbon $C_t$. $v$ = 0.1 denotes the time-varying rate of forcing $T_a(t)$, under which R-tipping will not occur in absence of noise. $D_{3}=0.5$ determines the magnitude of noise perturbations $dW_{3 t}$. For the remaining parameters, their definitions and chosen values in this study are provided in SI Appendix, Tab. S1.

The stochastic differential equations were numerically integrated by an Euler-Maruyama scheme to obtain time series data. The integration step for the Saddle–node system and Bautin system is 0.01, while for the Compost-bomb system, it is 0.1. For each of the three systems, we repeated the simulations 200,000 runs, each run incorporating a different realization of white noise. This enabled ensemble simulations, resulting in 200,000 distinct time series for subsequent analysis. 

\subsection*{Data preprocessing}

We initially selected time series from the ensemble simulation dataset that displayed R-tipping. Their shared characteristic is a sudden increase in time series values within a very few time steps. Conversely, the ensemble time series without R-tipping all evolve following similar trajectories, whereby a time series envelope for the non-tipping scenario can be identified (see e.g. the grey shaded area in Fig. 2A). To determine the occurrence time of R-tipping for each time series, we identified it as the moment when a time series surpasses the envelope of non-tipping time series. For each time series that manifests R-tipping, their R-tipping times vary widely. Accordingly, we truncated the time series at the point in time at which R-tipping occurs, and marked that moment as time 0. The time series segment before this point was retained, allowing us to explore common patterns preceding the occurrence of R-tipping. For each dataset generated for the three simulated systems, we categorized them into group A and group B, corresponding to R-tipping and non-tipping scenarios. In case of the Compost-bomb system, all time series exhibit a common linear trend with the same slope (Fig. 2C). Consequently, we removed this linear trend from each time series before proceeding with subsequent truncation and grouping.

\subsection*{Calculating critical slowing down indicators}

Autocorrelation and variance calculations for all time series followed the same procedure. Each time series underwent nonlinear detrending using a running mean with a window size of 100 time steps. Subsequently, the CSD indicators were computed within a sliding window of 120 time steps, in which the variance and autocorrelation are calculated using the standard way \cite{scheffer2009early, boers2022theoretical}. We experimented with different sizes for the sliding window, ranging from 60 to 200 time steps. The resulting CSD indicators did not show significant divergence, and the insights for the inferences remained consistent. 

\subsection*{Configurations for deep learning}

We utilized a CNN-based framework to construct our DL model for predicting R-tipping. As illustrated in SI Appendix, Fig. S1, the model is composed of two one-dimensional CNN layers and one fully-connected network layer, with average pooling procedures connecting them. The models were implemented using Pytorch 1.12.0. The convolution kernel sizes and filters were uniformly set as 3 and 64. The Cross Entropy loss function and Stochastic Gradient Descent optimizer were adopted, and the learning rate was set to 0.01. When a time series segment is input into the DL model, the sliding convolution process extracts local information from the data, while the fully-connected network can handle the global information of the data, through which we expect that the DL model can discern predictive information for R-tipping from the data. We provide detailed descriptions of the tasks for each model during the analysis of the results. For training each model, we set the numbers of samples in the training dataset and the testing dataset as 108,000 and 12,000 respectively. 

Prediction accuracy of the DL model is determined by calculating the ratio between correctly predicted cases and the total number of cases of R-tipping and non-tipping groups. A prediction case is considered correct when the predicted R-tipping probability of a R-tipping case is higher than 50\%, or that of a non-tipping case is lower than 50\%.

We employ layer-wise relevance propagation (LRP) to interprete and elucidate the decision-making process of or DL framework in classifying time series into tipping and non-tipping cases. The principle underlying LRP involves the backward distribution of the prediction score through the network's layers, back to the input features. Relevance scores, represented as LRP scores in Fig. 4, are assigned to each neuron and input feature based on their contribution to the final prediction. The algorithm traces and distributes the relevance of the output back to the input features while respecting the network's learned weights and activations. The process of relevance propagation in LRP is mathematically intricate, involving the application of rules and formulas to distribute relevance backward through the layers. Here we do not present its mathematical content. More detailed information on the LRP algorithm can be found in existing literature \cite{bach2015pixel, montavon2019layer}.

\bmhead{Data availability}
All data used in this study, together with the code for simulating the Saddle-node system, Bautin system and  Compost-bomb system, will be made availabe on Github after this manuscript is published: https://github.com/yhuangDLClimate/predict-rate-induced-tipping.

\bmhead{Code availability}
 The Matlab code for processing and analysing the data, together with the PyTorch code for implementing the deep learning model, will be made available on Github after this manuscript is published: https://github.com/yhuangDLClimate/predict-rate-induced-tipping.

\bmhead{Acknowledgements}
Y.H. acknowledges Alexander von Humboldt Foundation for Humboldt Research Fellowship. N.B. and S.B. acknowledge funding by the Volkswagen foundation. NB acknowledges further funding by the European Union’s Horizon 2020 research and innovation programme under the Marie
Sklodowska-Curie grant agreement No.956170. This is ClimTip contribution \#X; the ClimTip project has received funding from the European Union's Horizon Europe research and innovation programme under grant agreement No. 101137601. The authors acknowledge the support of high performance computing resource by the Leibniz Supercomputing Centre. For the purpose of open access, the authors have applied a Creative Commons Attribution (CC BY) licence to any Author Accepted Manuscript version arising from this submission. An alternative statement with the same intended outcome can be used if required by specific funders.

\bmhead{Competing interests}
The authors declare no competing interests.


\bibliography{sn-bibliography}

\clearpage
\begin{figure}
\centering
\includegraphics[width=.9\linewidth]{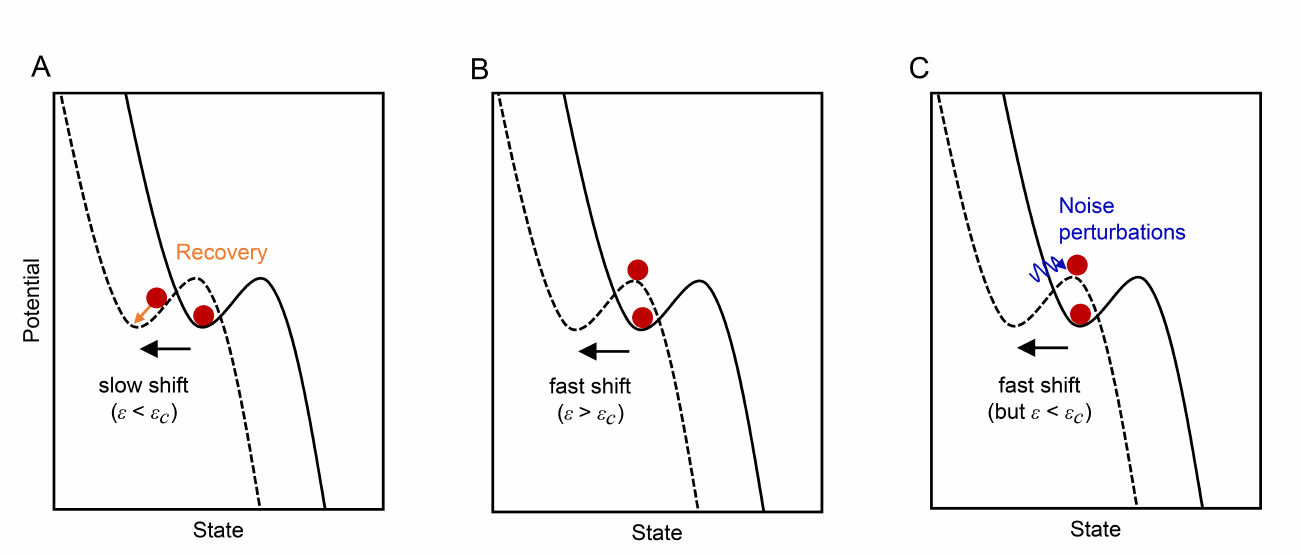}
\caption{Diagram depicting R-tipping through visualization of a system represented as a particle in a basin of attraction. (A) The comparison of basins of attraction before (black solid line) and after (black dash line) undergoing a shift due to the change in environmental forcing. When the time-varying rate of environmental forcing change (i.e., forcing rate $\epsilon$) is much slower than the critical threshold ($\epsilon_c$), the particle near to the basin (its quasi-equilibrium state) can recover timely and recover to the equilibrium state (i.e., the stable fixed point); in other words, the forcing rate is sufficiently low for the system to be able to track the basin of attraction associated with that equilibrium. (B) When the forcing rate is faster than the critical threshold, the particle cannot track the basin of attraction anymore, and R-tipping occurs. (C) When the forcing rate is rapid but stays slightly below the critical threshold, but the system additionally experiences slight noise perturbations, the particle will leave the basin of attraction for some noise realisations, but not for others, establishing a mixture of rate- and noise-induced tipping. }
\label{fig:figure1}
\end{figure}

\clearpage
 \begin{figure*}[htbp]
\centering  
\includegraphics[width=.32\linewidth]{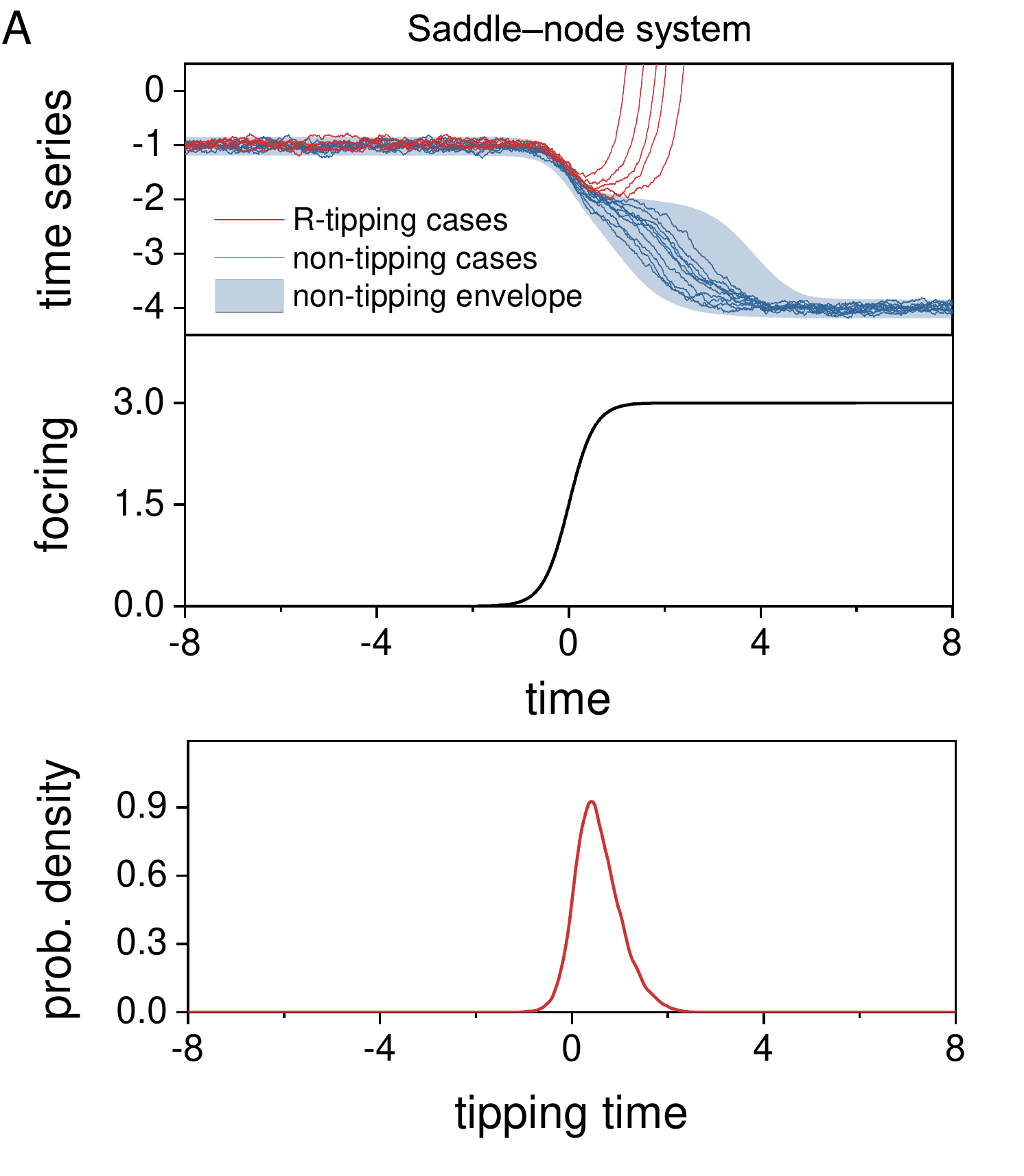}
\includegraphics[width=.32\linewidth]{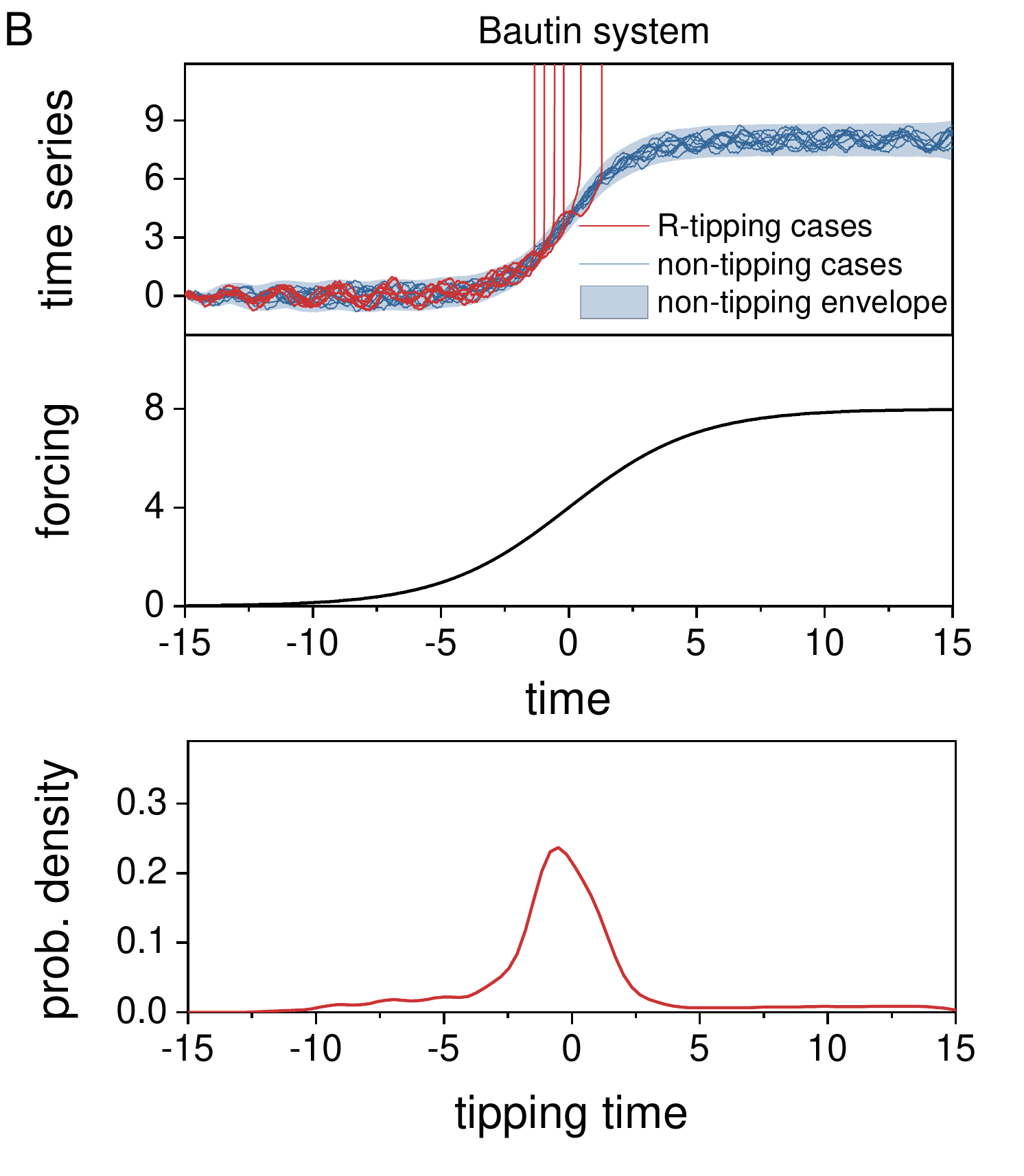}
\includegraphics[width=.32\linewidth]{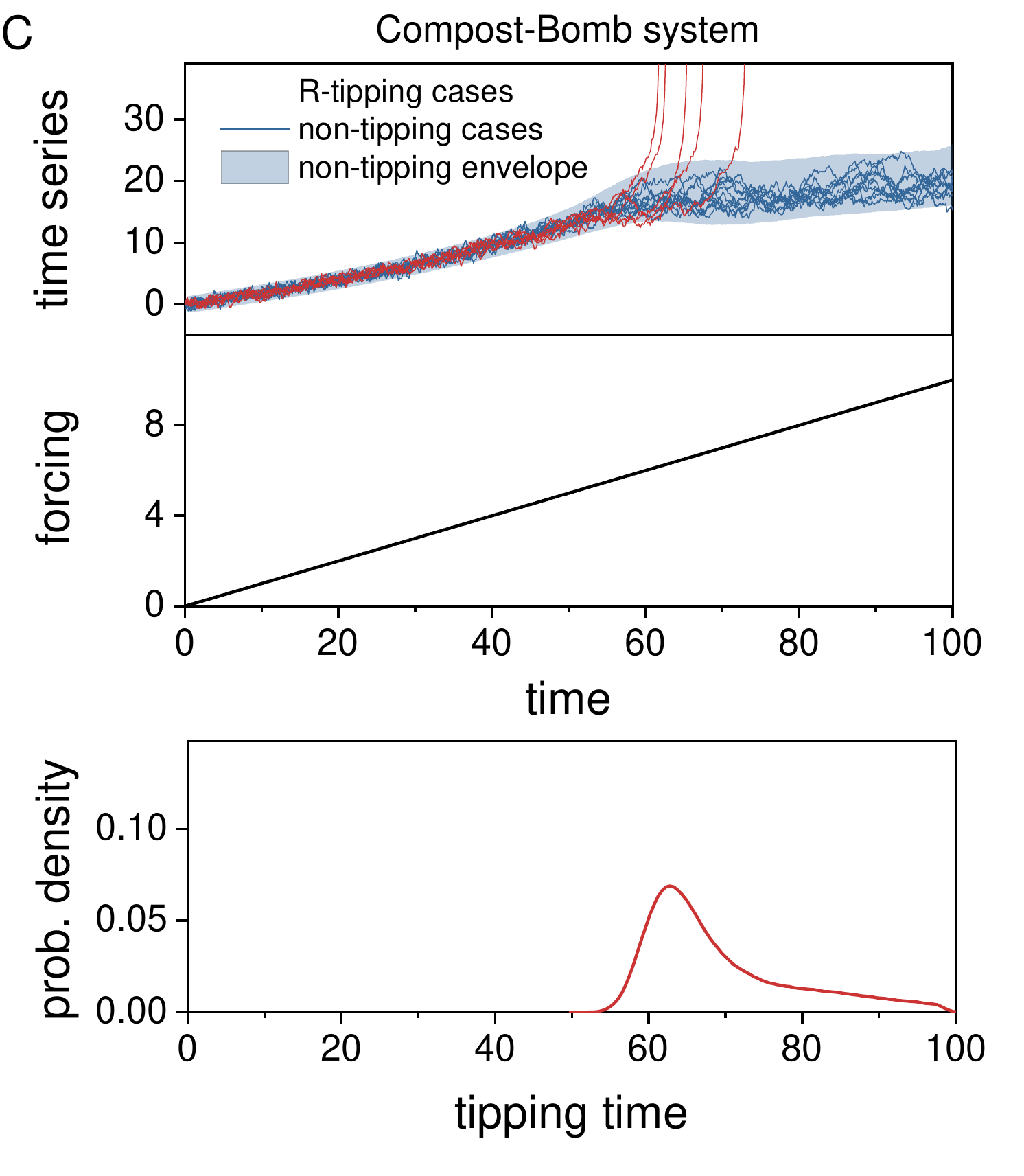}
\caption{Ensemble simulation results for prototype R-tipping systems. Ensemble simulations are conducted on the Saddle–node system (A), Bautin system (B), and Compost-bomb system (C), respectively. For each simulated system, the same time-varying forcing parameter but different noise perturbations are used for the simulations. Top panels: simulated time series that do not tip (blue) and that exhibit R-tipping (red). Middle panels: The time-varying forcing parameter. Bottom panels: The probability density of the observed occurrence time of R-tipping. Blue shading area denotes the 99\% confidence intervals of the ensemble realizations which do not manifest R-tipping. }
\end{figure*}

\clearpage
\begin{figure*}[htbp]
\centering  
\includegraphics[width=.32\linewidth]{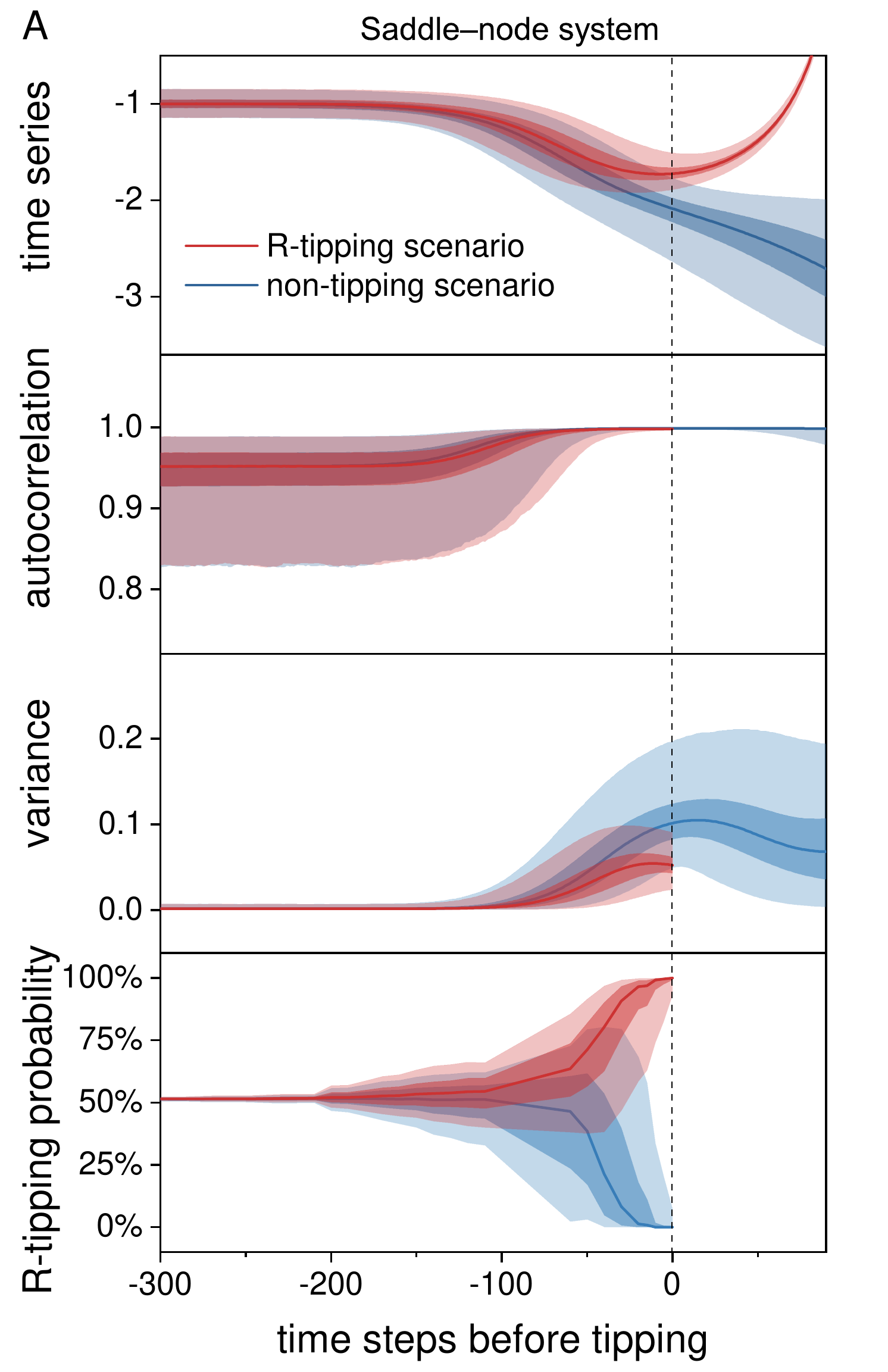}
\includegraphics[width=.32\linewidth]{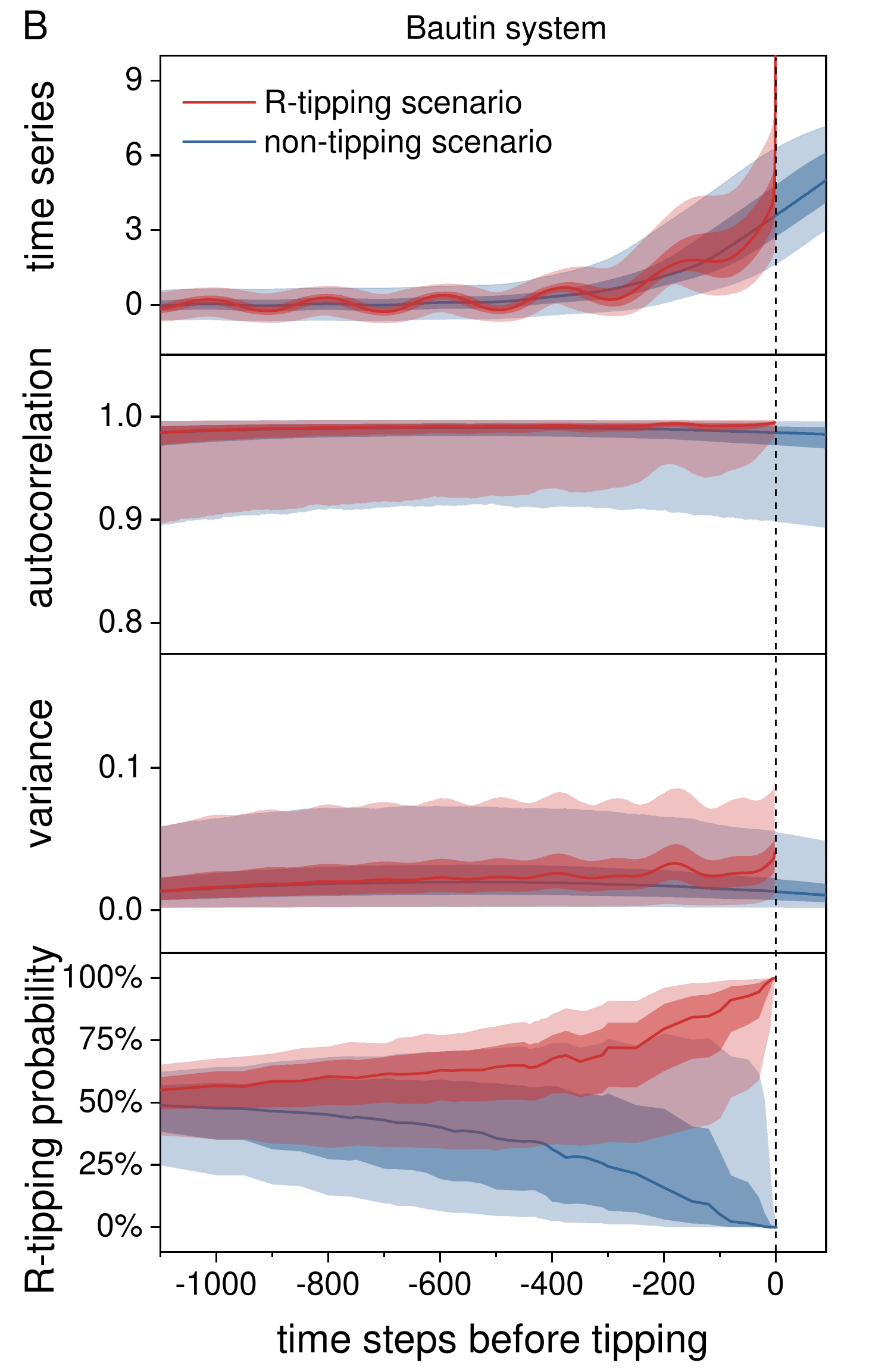}
\includegraphics[width=.32\linewidth]{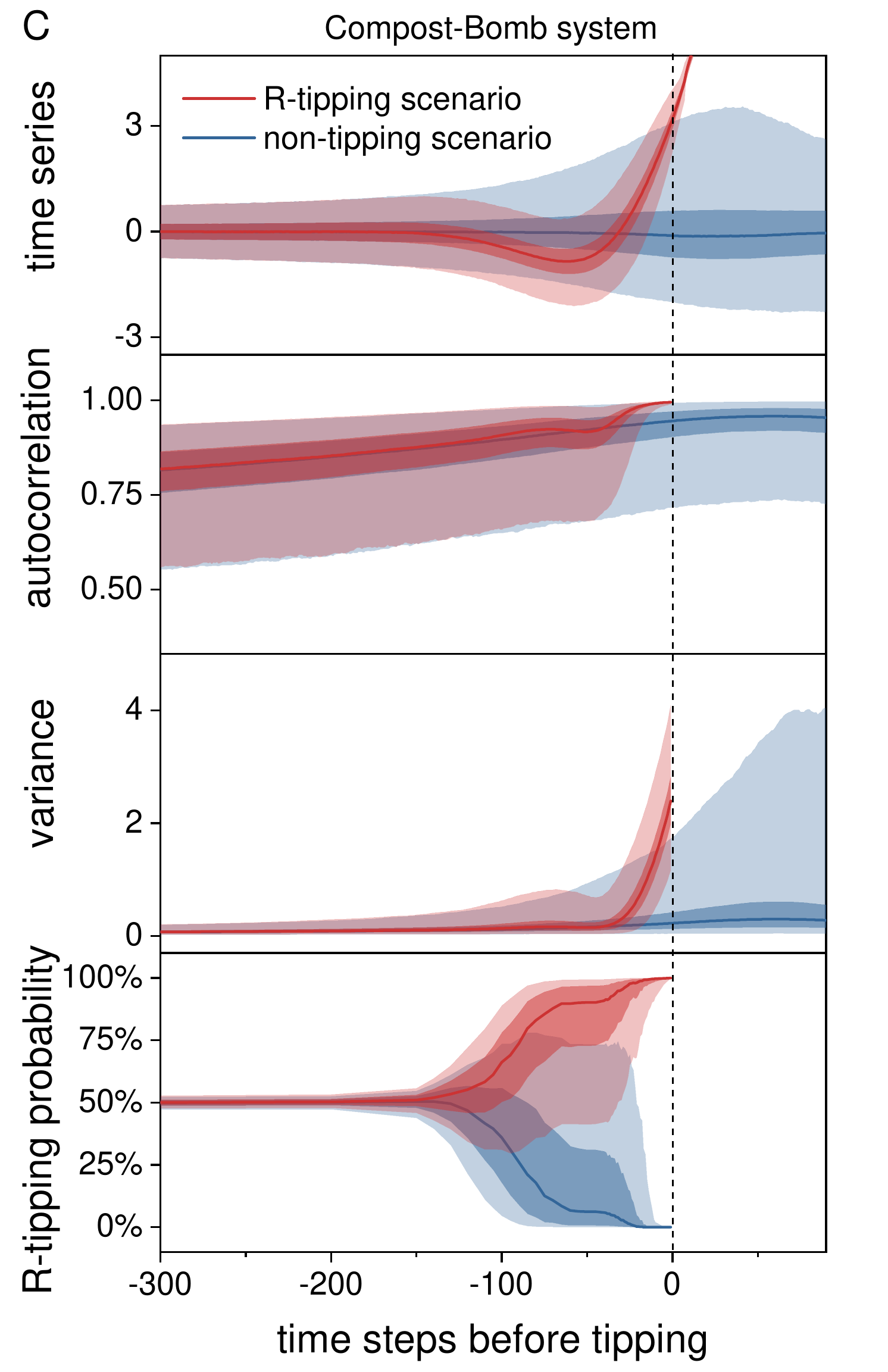}
\caption{Deep-learning based prediction of R-tipping for the paradigmatic systems such as the Saddle–node system (A), Bautin system (B), and Compost-bomb system (C), and comparison to classical early-warning indicators. Top panel: statistics of simulated time series for indicating the time-varying system states. Upper middle panel: estimated autocorrelation evolution. Lower middle panel: estimated variance. Bottom panel: DL-derived R-tipping probabilities as functions of time. The red solid lines represents the composite mean values of time series within the class exhibiting R-tipping, while light red and red shading areas depict the 99\% and 75\% confidence intervals, respectively. The non-tipping scenarios are indicated by blue correspondingly. The unit on the time axis adopts the time step used in numerical integration (see Materials and Methods). }
\end{figure*}

\clearpage
\begin{figure*}[htbp]
\centering  
\includegraphics[width=.75\linewidth]{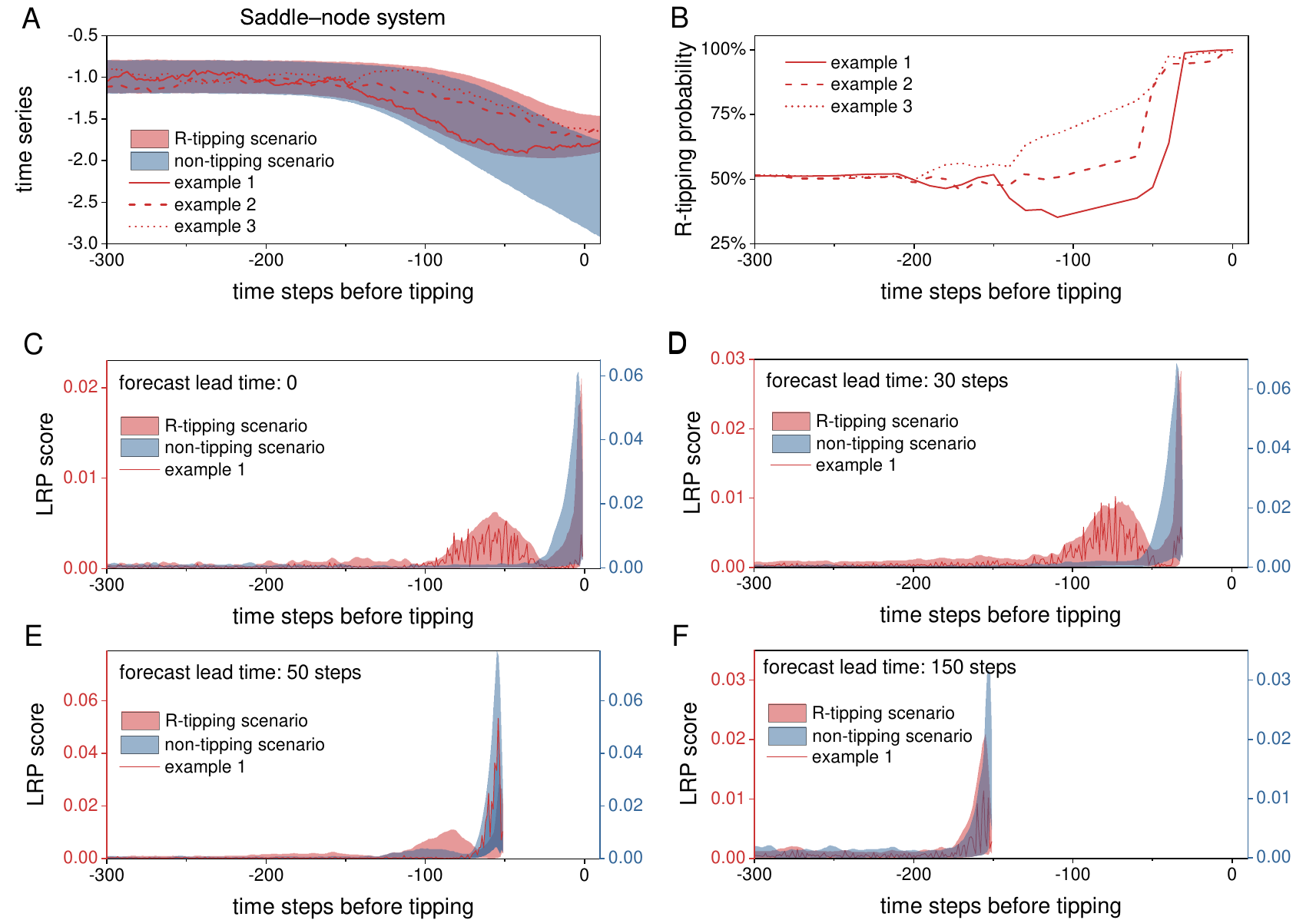}
\caption{Saliency maps to interpret the fingerprint features that the trained DL models for predicting R-tipping have extracted. (A) Three example time series (from the Saddle–node system) approaching R-tipping (red solid, dashed and doted lines), and the 99\% confidence intervals of R-tipping and non-tipping scenarios (red and blue shading areas). (B) The predicted R-tipping probabilities for the three example time series. (C) LRP scores for a lead time equal to 0 time steps, indicating the relative importance of each individual time series point for guiding the prediction. Blue solid line shows the LRP scores for the time series of example 1. Red and blue shading areas represent the 99\% confidence intervals of the LRP scores for R-tipping and non-tipping scenarios, respectively. The left and right vertical axes denote the LRP values of R-tipping and non-tipping scenarios, respectively. (D), (E), and (F) mirror the configuration of (C), corresponding to lead times of 30, 50 and 150 time steps, respectively. }
\end{figure*}

\clearpage
\begin{figure}
\centering
\includegraphics[width=0.31\textwidth]{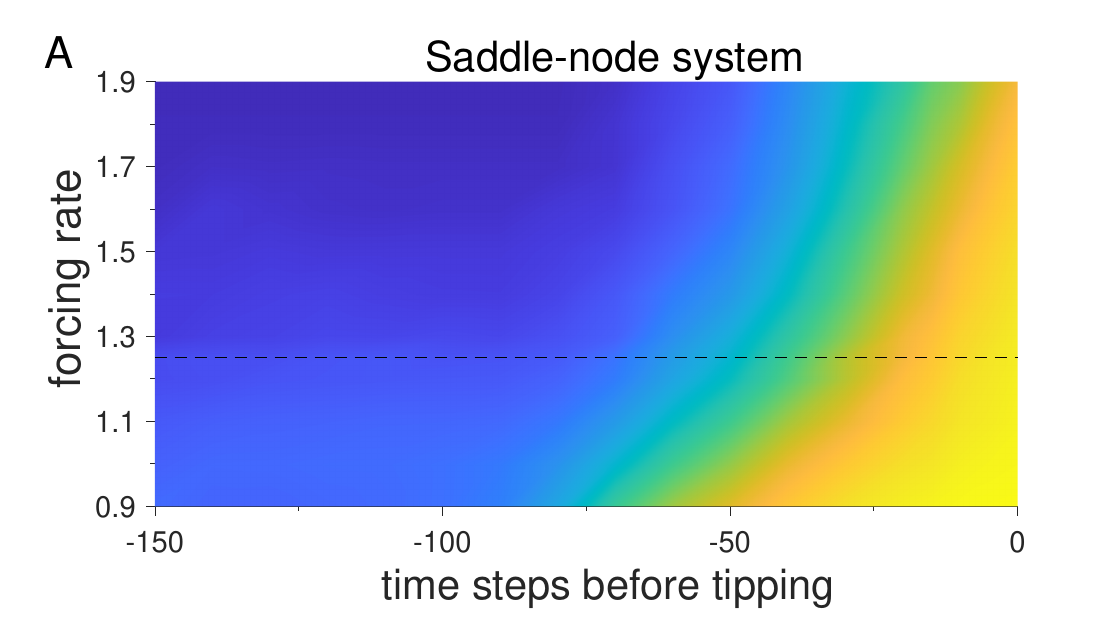}
\centering
\includegraphics[width=0.31\textwidth]{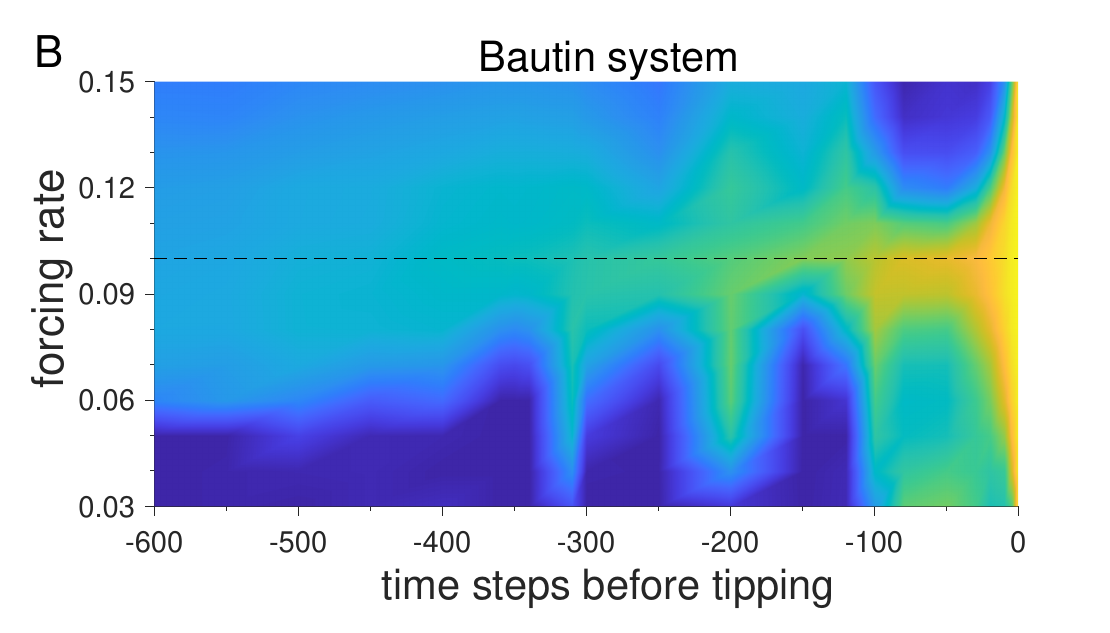}
\centering
\includegraphics[width=0.31\textwidth]{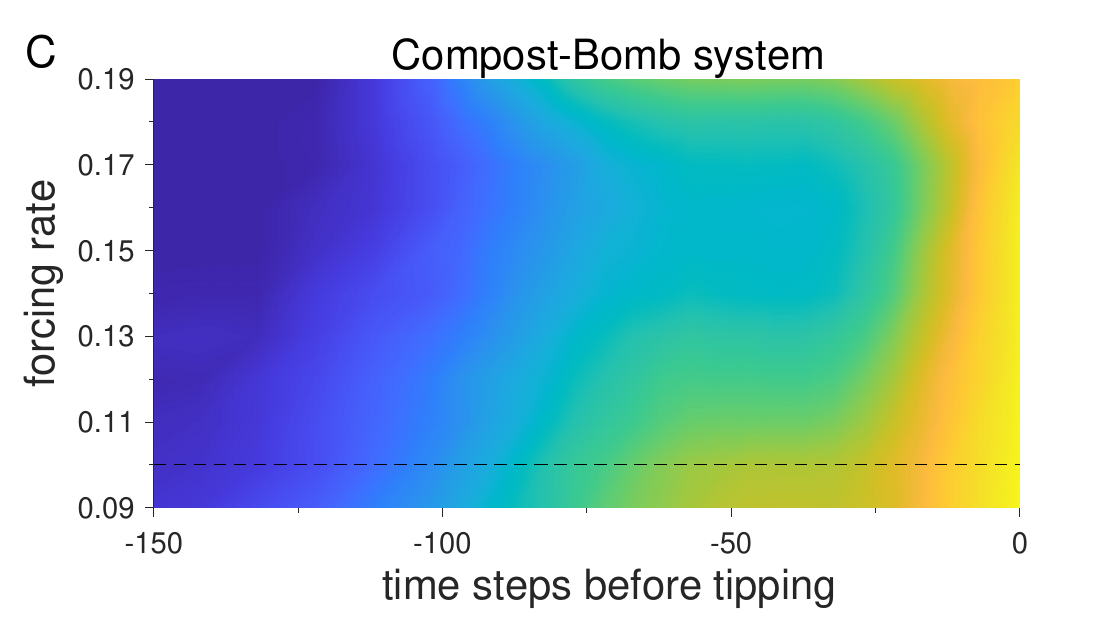}
\qquad
\centering
\includegraphics[width=0.31\textwidth]{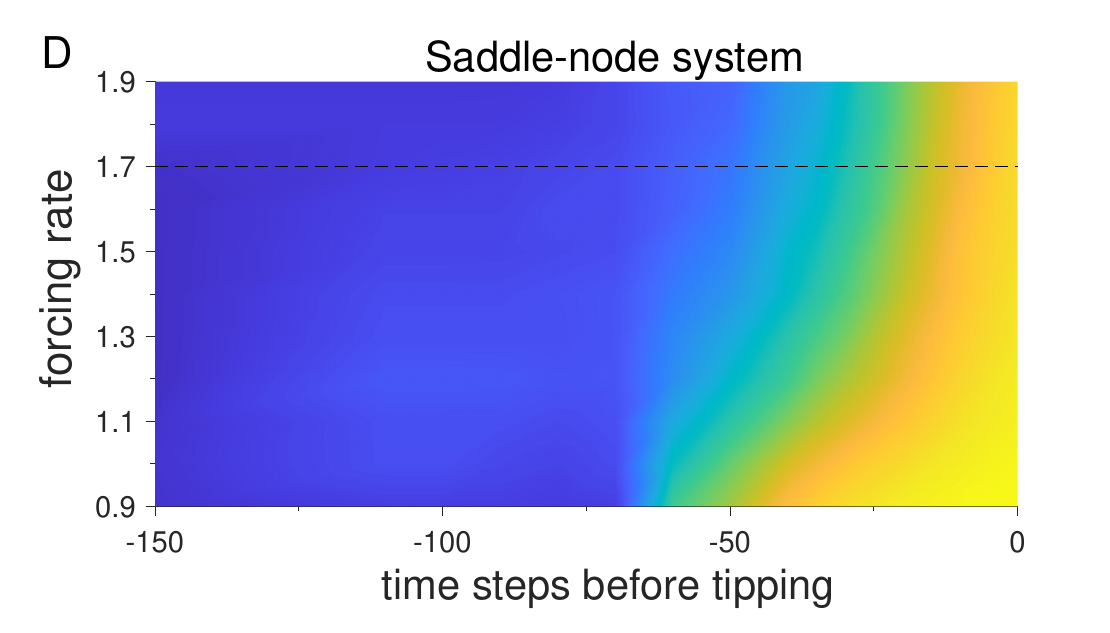}
\centering
\includegraphics[width=0.31\textwidth]{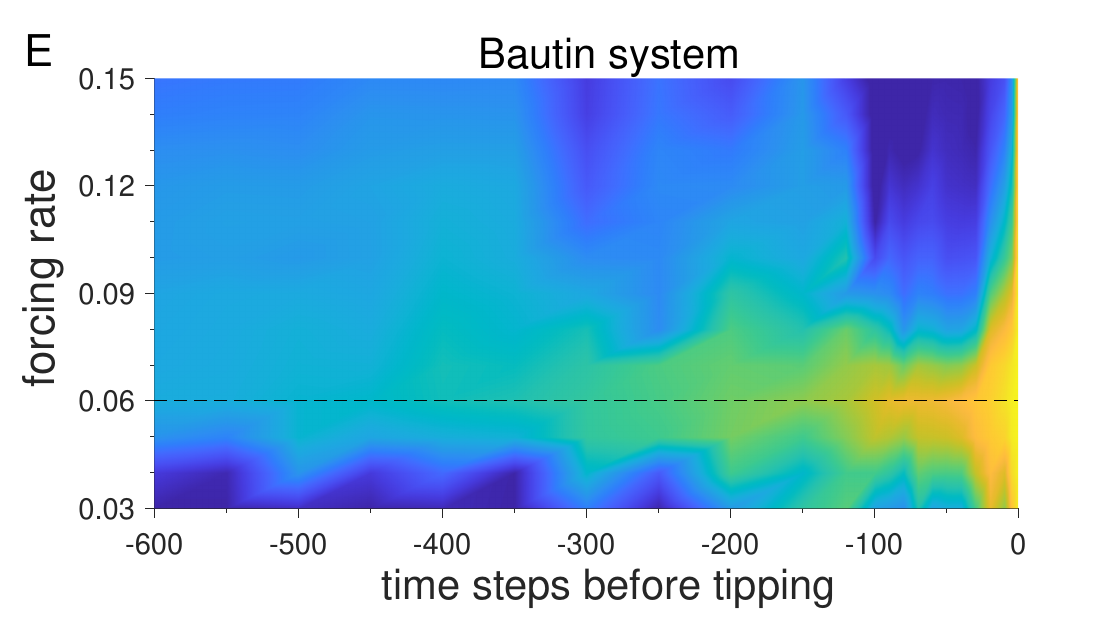}
\centering
\includegraphics[width=0.31\textwidth]{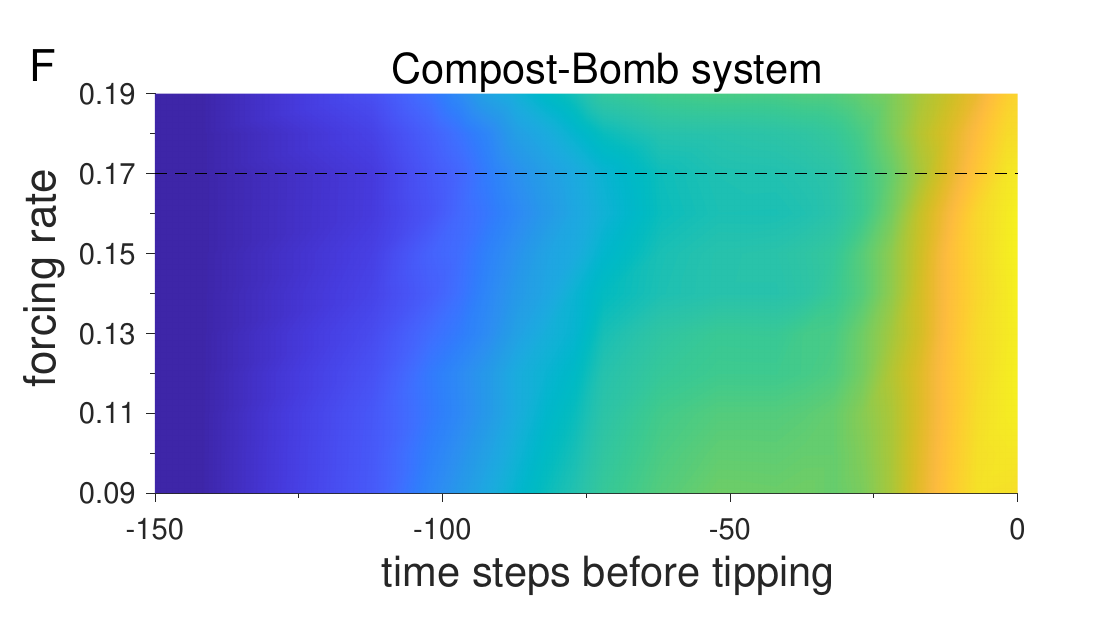}
\qquad
\centering
\includegraphics[width=0.31\textwidth]{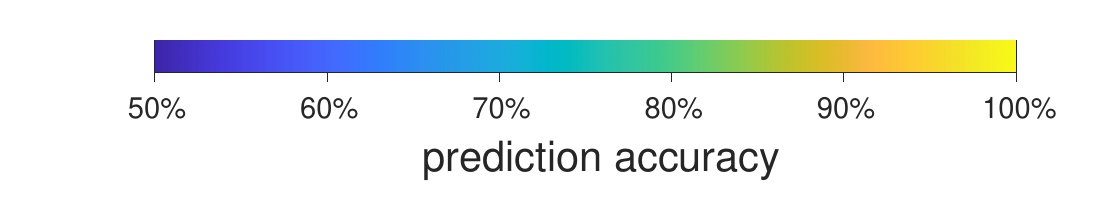}
\caption{Prediction accuracy of DL models for R-tipping with out-of-sample forcing rates. For the Saddle–node system, DL models were trained on time series with specific forcing rates $\epsilon=1.25$ (A) and $\epsilon=1.7$ (D), respectively, and subsequently used to predict R-tipping cases with previously unseen forcing rates, and the prediction accuracy as a function of forcing rate and forecast lead time is shown. For the Bautin system, DL models were trained on forcing rates $r=1.0$ (B) and $r=0.6$ (E), respectively. For the Compost-bomb system, DL models were trained on forcing rates $v=0.1$ (C) and $v=0.17$ (F), respectively.}
\end{figure}

\end{document}


\clearpage
\section*{Supplementary Information Appendix}

\clearpage
\begin{figure}
\centering
\includegraphics[width=0.3\textwidth]{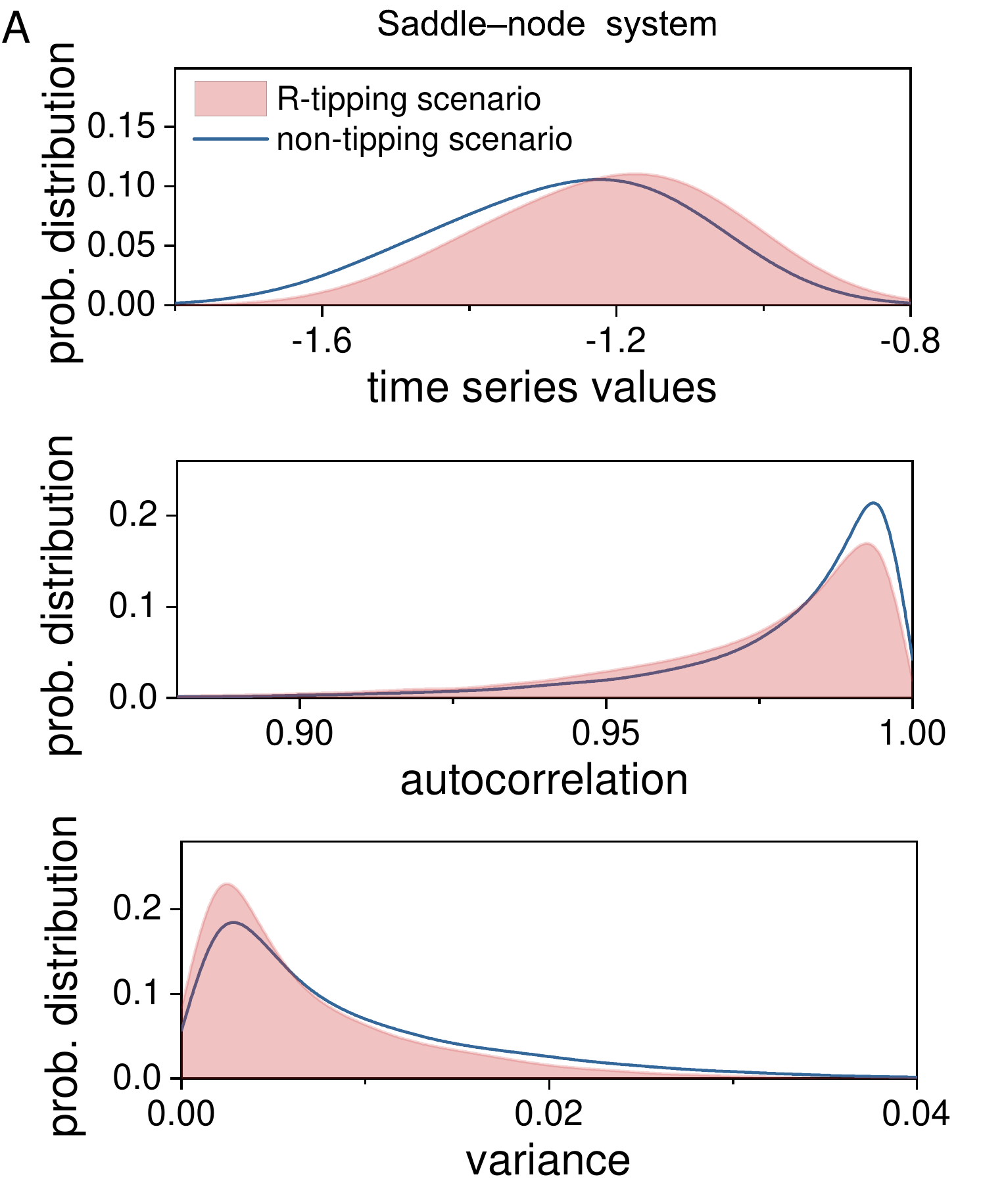}
\centering
\includegraphics[width=0.3\textwidth]{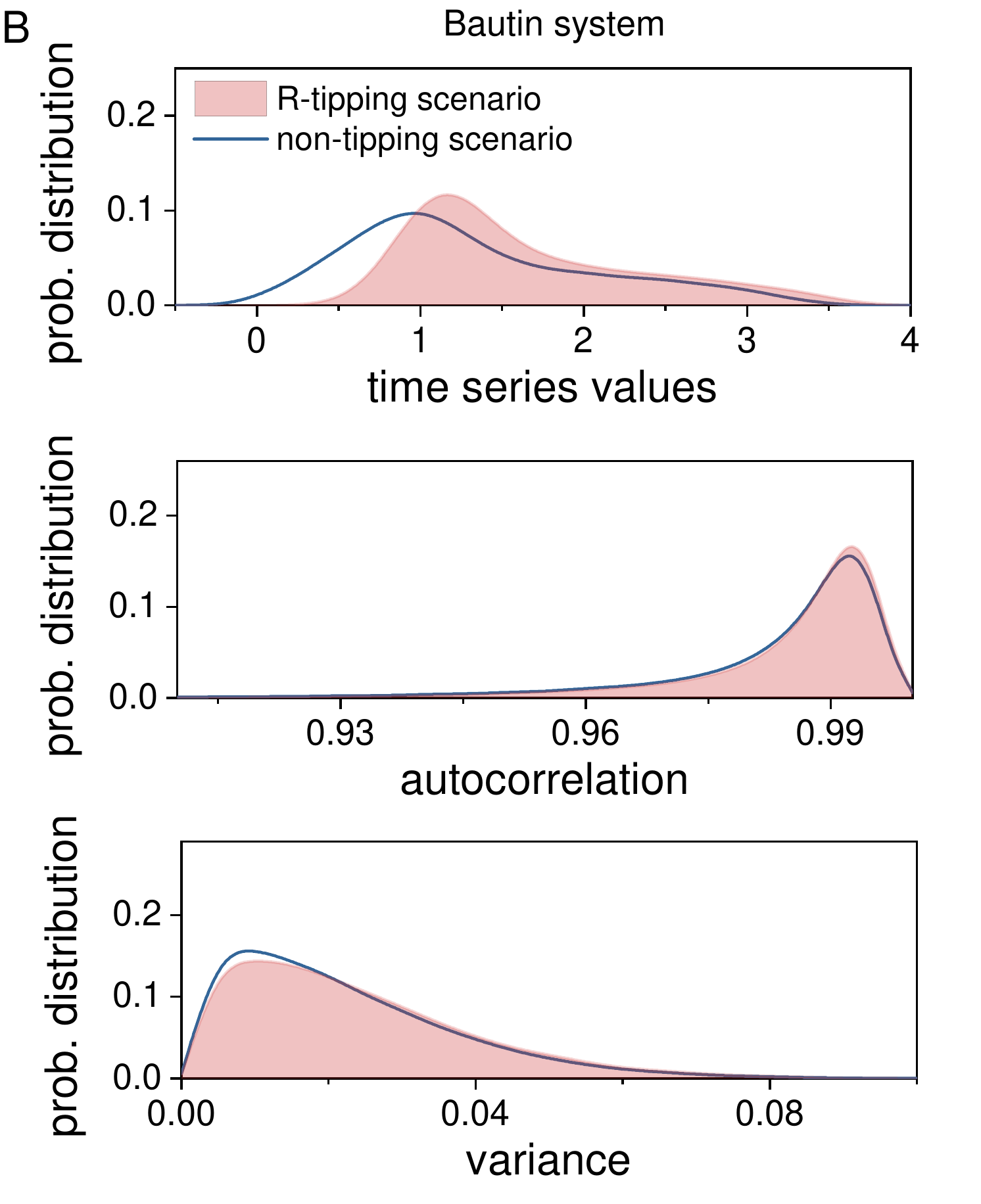}
\centering
\includegraphics[width=0.3\textwidth]{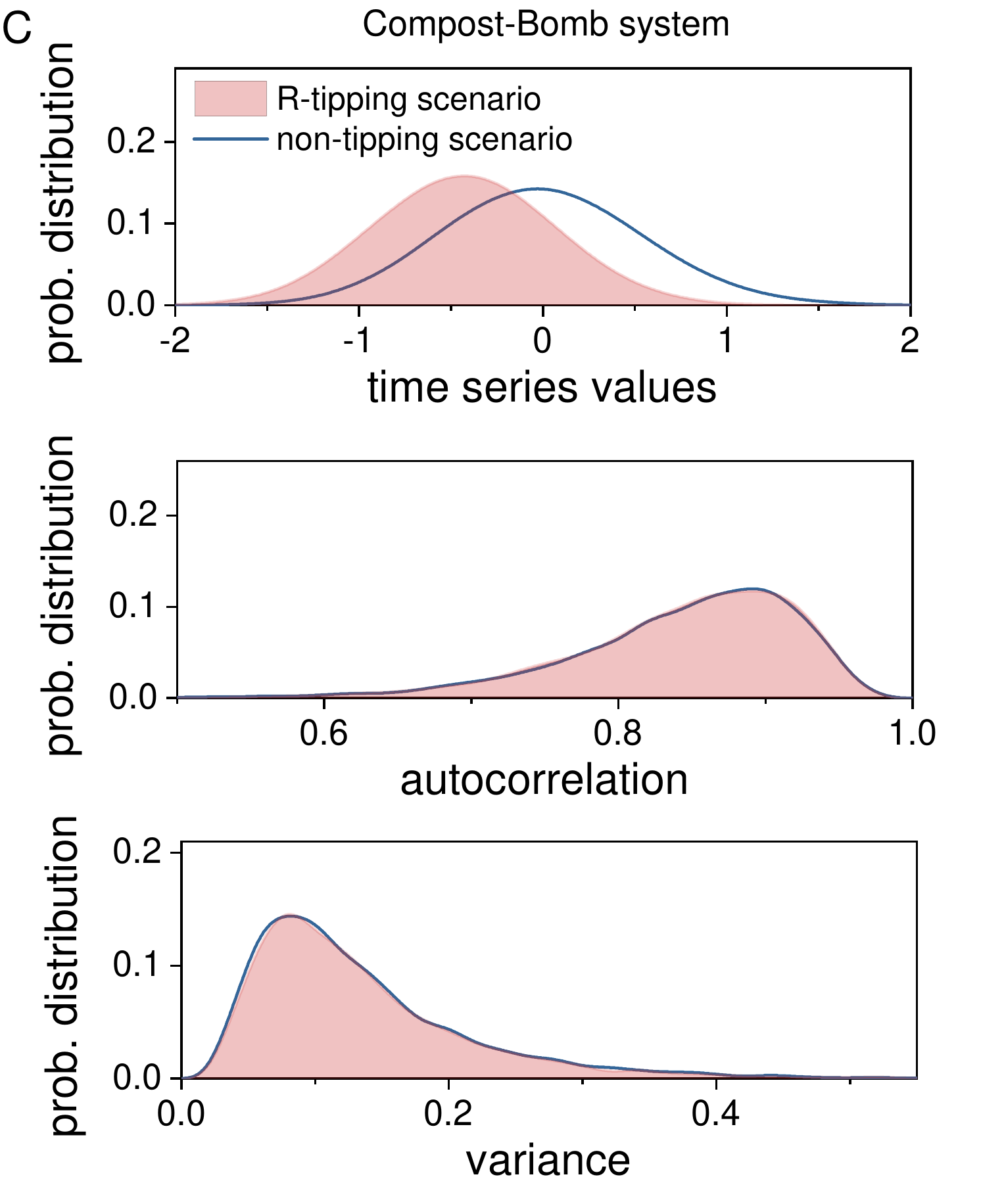}
\caption{Probability distributions of system states and CSD indicators leading up to the R-tipping occurrence, and comparison between R-tipping and non-tipping scenarios. Statistics analysis is conducted on Saddle–node system at 100 time steps leading to the R-tipping occurrence (A), on Bautin system at 200 time steps leading to the R-tipping occurrence (B), and on Compost-Bomb system at 100 time steps leading to the R-tipping occurrence (C), respectively. Top panels: probability distributions of time series values. Middle panels: autocorrelations. Bottom panels: variances.}
\end{figure}

\clearpage
\begin{figure}
\centering
\includegraphics[width=0.3\textwidth]{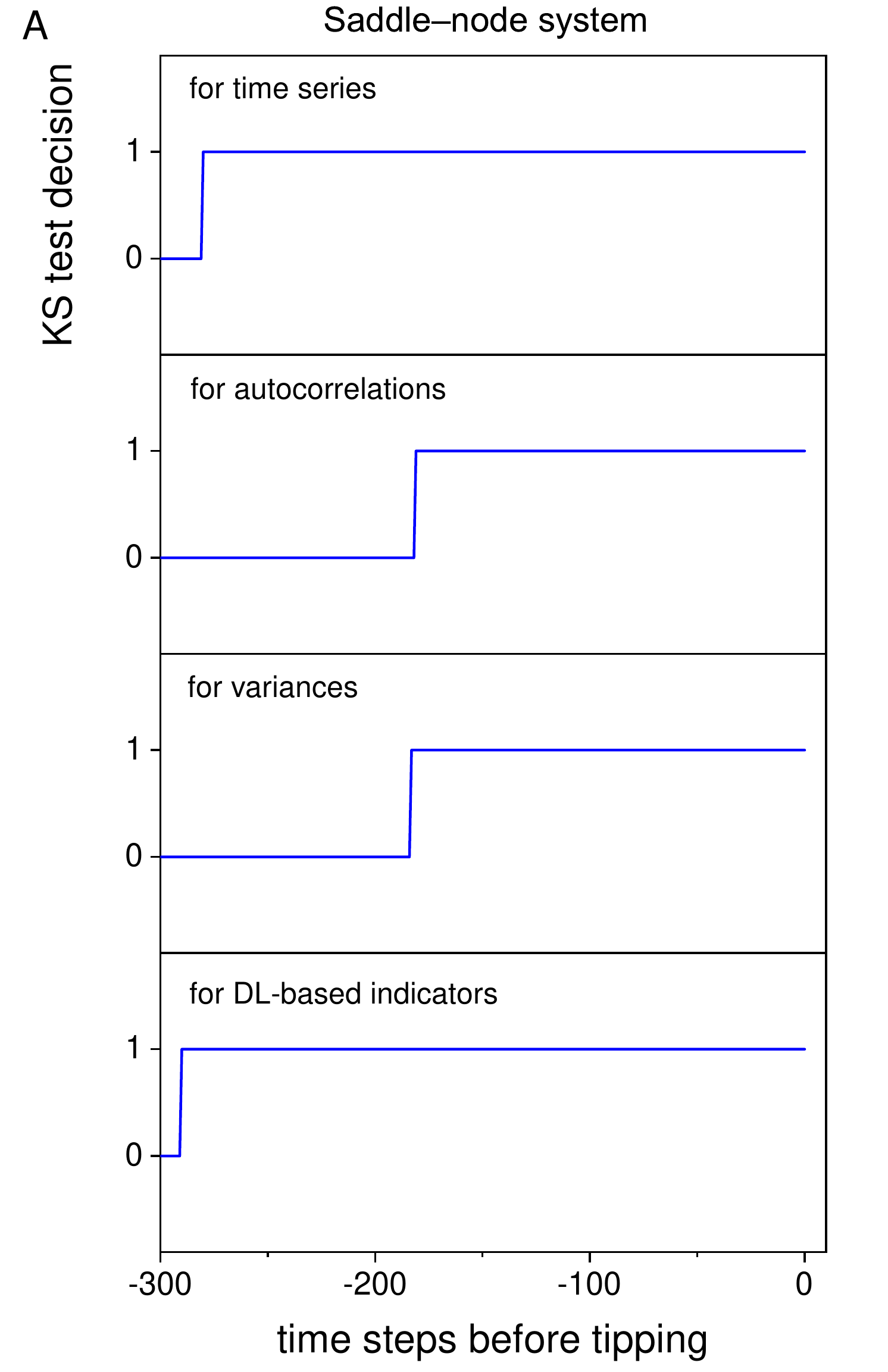}
\centering
\includegraphics[width=0.3\textwidth]{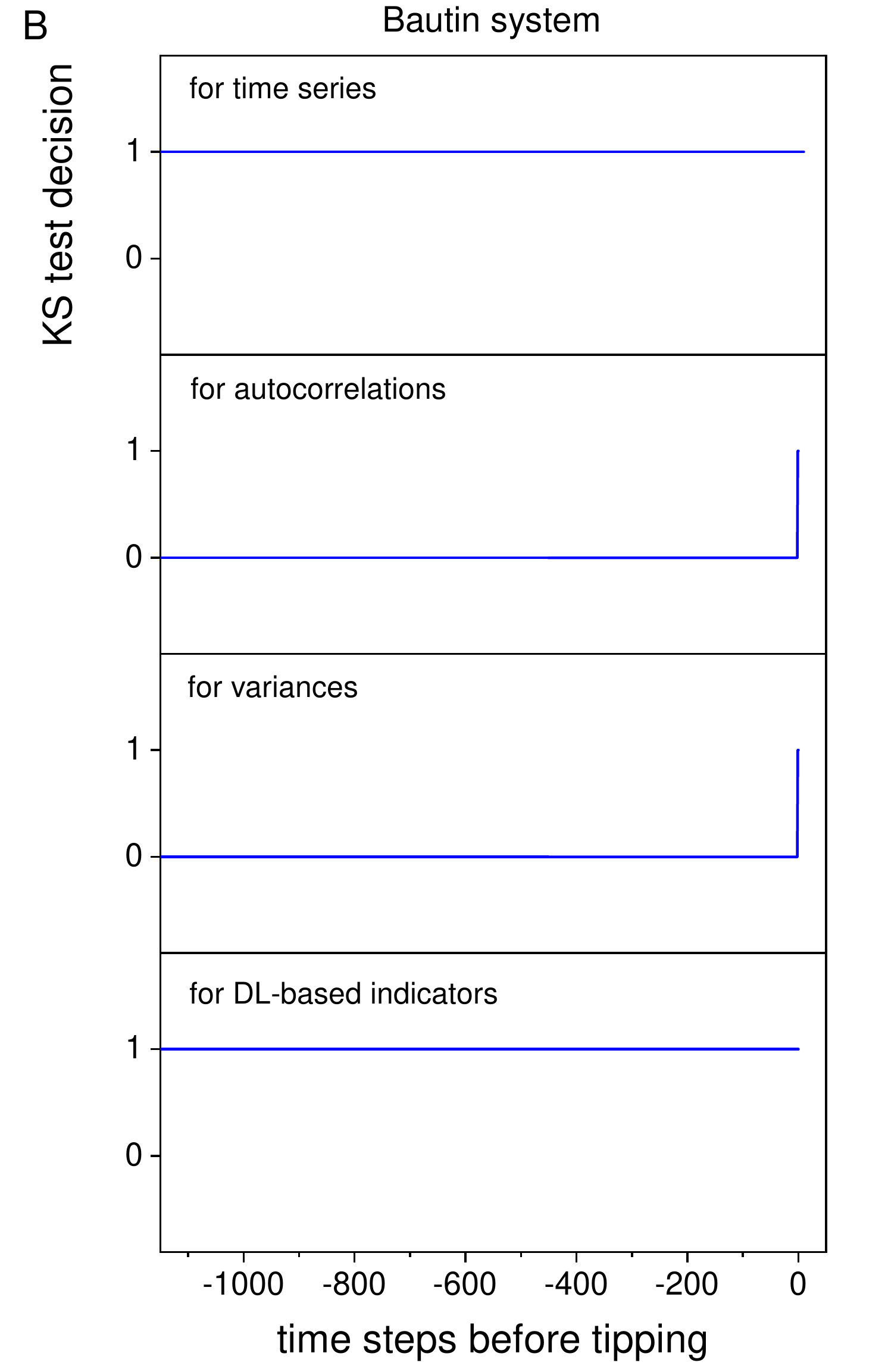}
\centering
\includegraphics[width=0.3\textwidth]{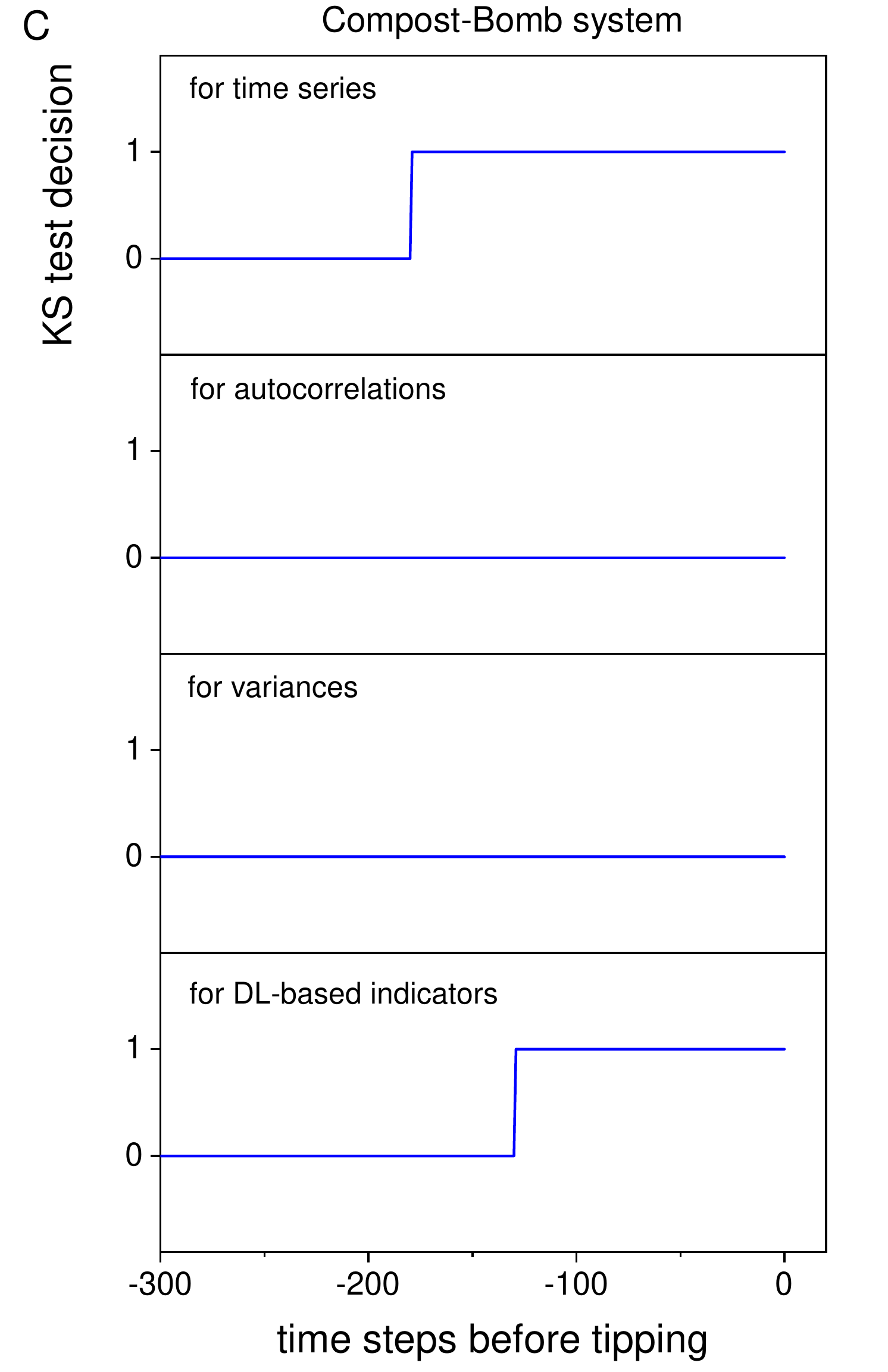}
\caption{The results of the two-sample Kolmogorov-Smirnov (KS) significance test are presented, for comparison on probability distributions of the system states, CSD indicators and DL-based R-tipping indicators between the R-tipping and non-tipping scenarios. A decision of "1" on the KS test indicates a significant difference in the probability distributions between the R-tipping and non-tipping scenarios, with a significance level of 0.01. Conversely, a "0" denotes that the observed difference is not statistically significant. Analysis is conducted on Saddle–node system (A), Bautin system (B), and Compost-Bomb system (C), respectively, across different leading time prior to R-tipping occurrence. Top panels: KS test results on time series values. Upper middle panels: autocorrelations. Lower middle panels: variances. Bottom panels: DL-based R-tipping probabilities. }
\end{figure}

\clearpage
\begin{figure}
\centering
\includegraphics[width=\textwidth]{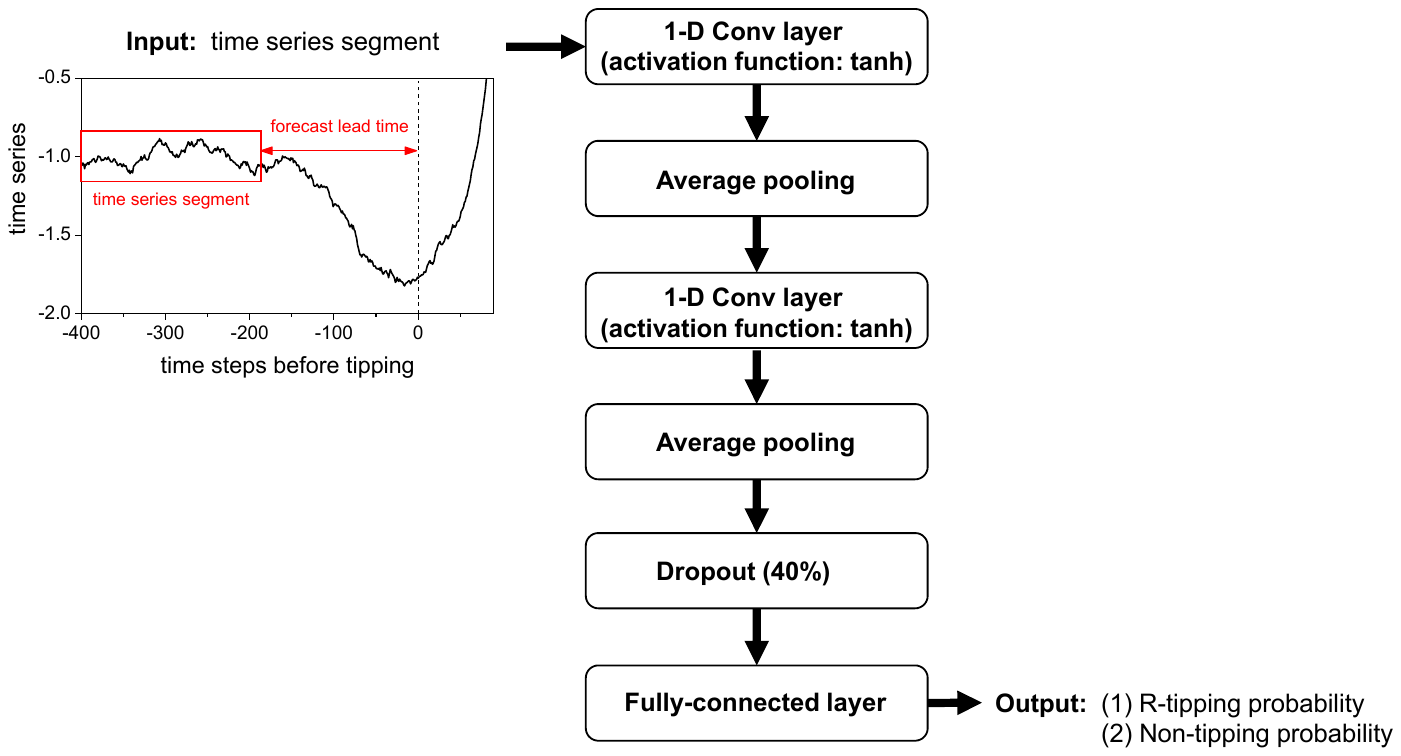}
\caption{Schematic illustrating the architecture and workflow of the DL model. For a certain lead time before the R-tipping occurrence, a time series segment is feed into the DL model, the binary outputs denote the probabilities of this time series segment being inferred as either an R-tipping scenario or a non-tipping scenario, respectively. The DL model consists of one-dimensional convolution neural network layer (1-D Conv), average pooling layer, and fully connected neural network layer. }
\end{figure}

\clearpage
\begin{figure}
\centering
\includegraphics[width=0.3\textwidth]{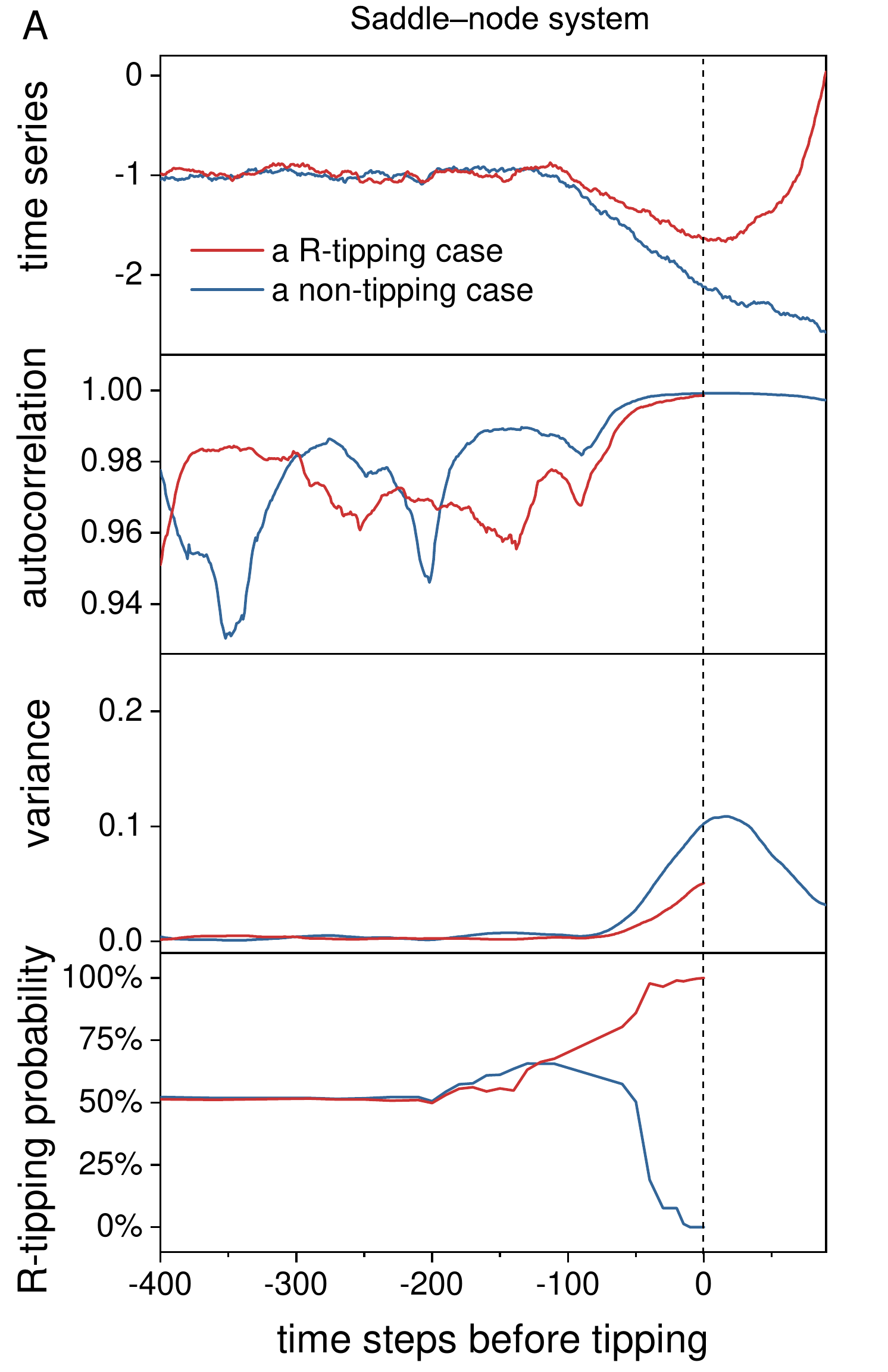}
\centering
\includegraphics[width=0.3\textwidth]{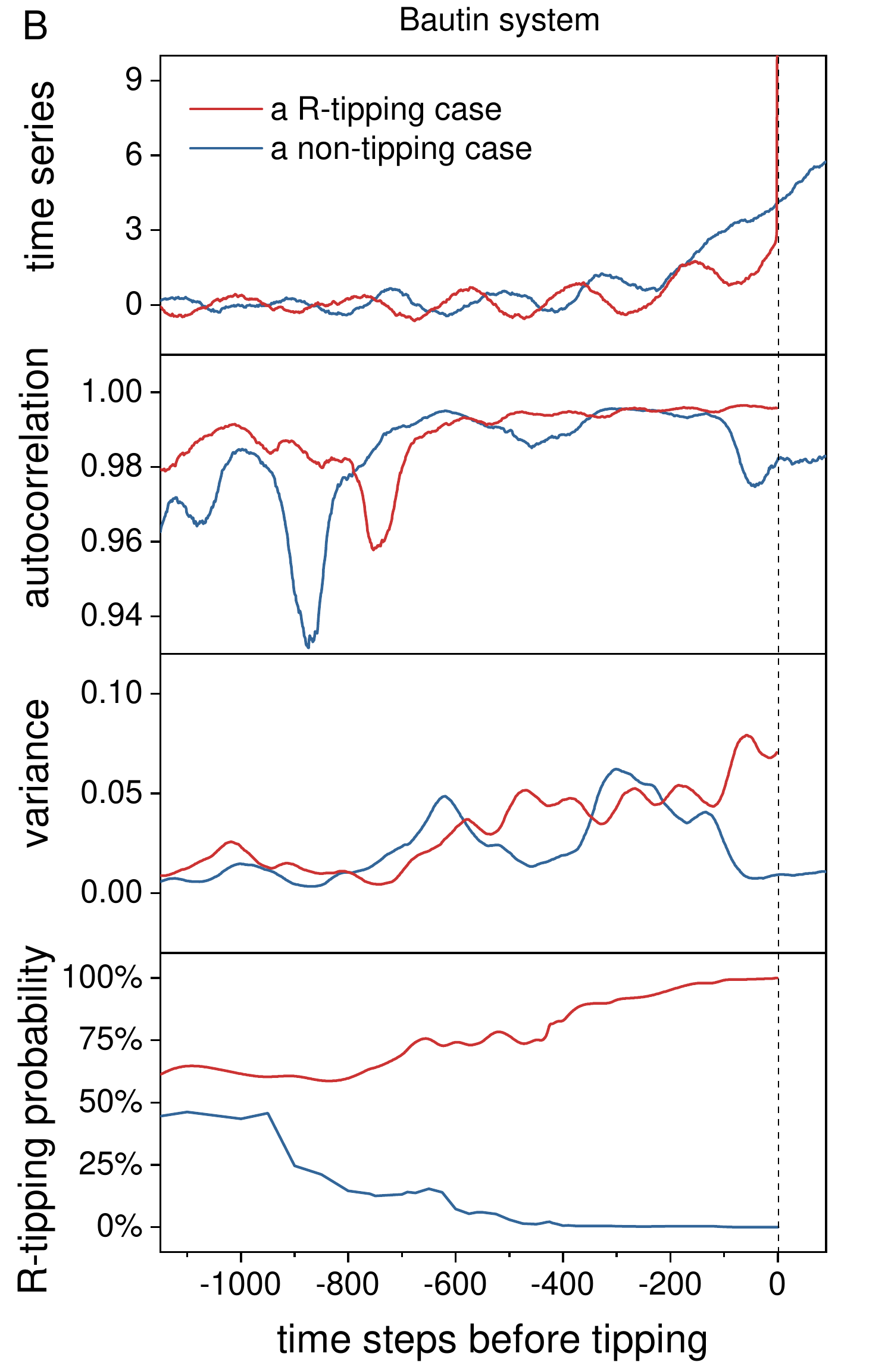}
\centering
\includegraphics[width=0.3\textwidth]{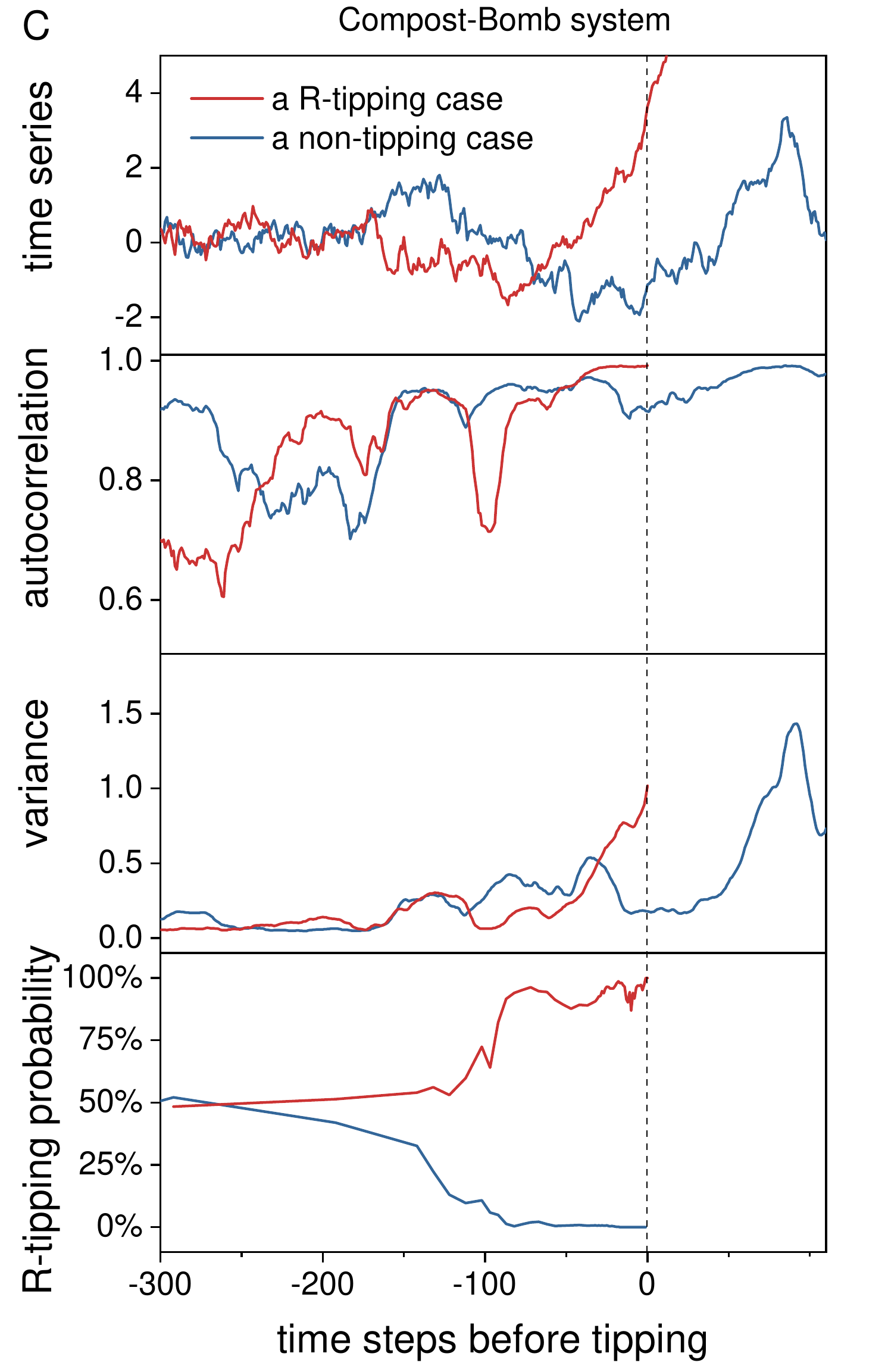}
\caption{Deep-learning based prediction of R-tipping for the paradigmatic systems such as the Saddle–node system (A), Bautin system (B), and Compost-Bomb system (C), and comparison to classical early-warning indicators. Top panel: simulated time series
for indicating the time-varying system states. Upper middle panel: estimated autocorrelation evolution.
Lower middle panel: estimated variance. Bottom panel: DL-derived R-tipping probabilities as
functions of time. The blue and black solid lines represents example time series within the R-tipping and non-tipping scenarios, respectively.
The unit on the time axis adopts the time step used in numerical integrations (see Materials and
Methods).}
\end{figure}

\clearpage
\begin{figure}
\centering
\includegraphics[width=0.3\textwidth]{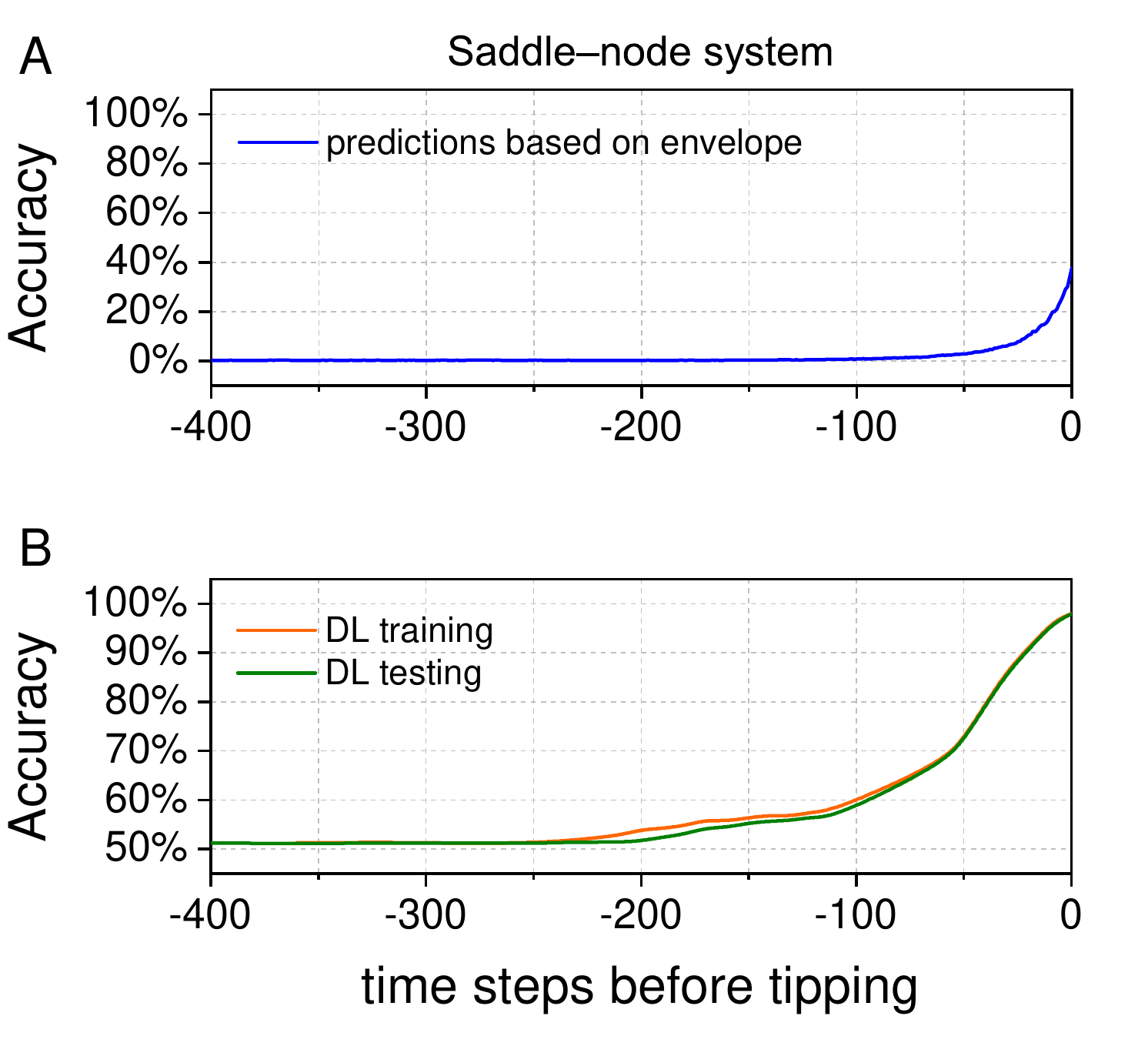}
\centering
\includegraphics[width=0.3\textwidth]{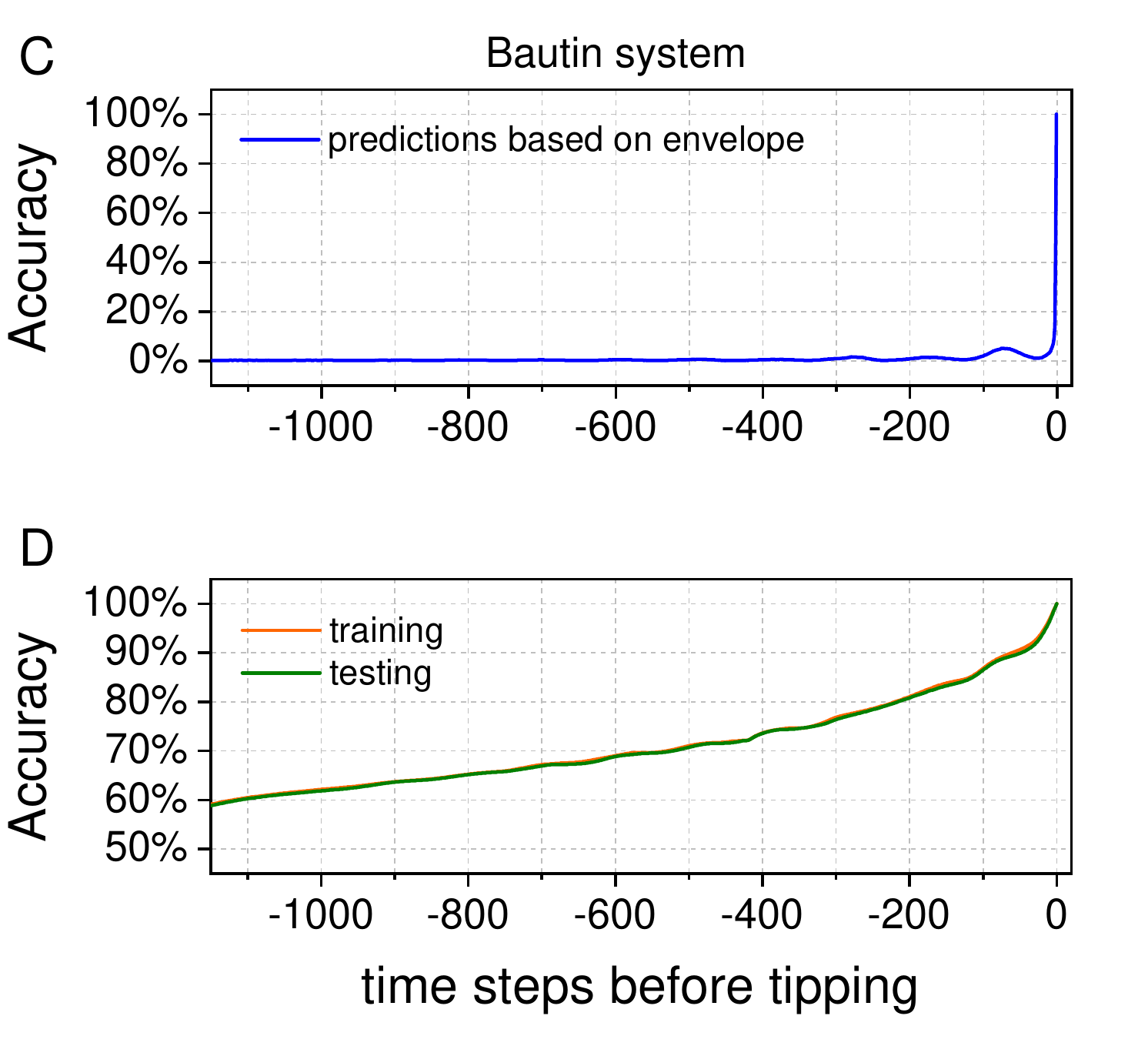}
\centering
\includegraphics[width=0.3\textwidth]{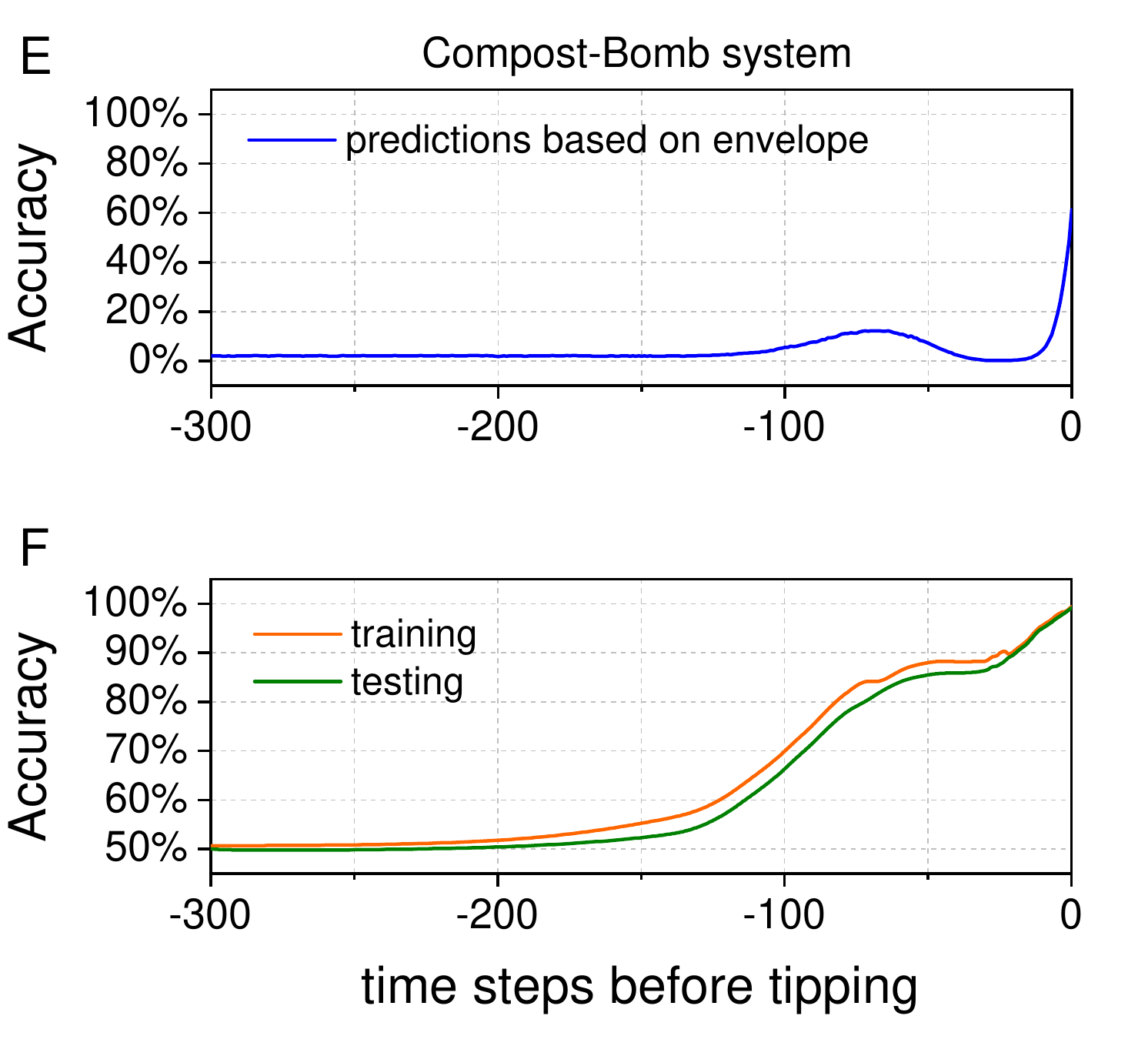}
\caption{Comparison of prediction accuracy between the time series envelope method and the DL model. Prediction accuracy of the time series envelope method is by calculating the ratio of group-A cases that fall outside the 99\% confidence interval of group B at each forecast lead time. Prediction accuracy of the DL model is by calculating the ratio between correct predicted cases and the total cases of group A and group B. A prediction case is considered correct when the predicted R-tipping probability of a time series in Group A is higher than 50\%, or that in Group B is lower than 50\%. (A) and (B): upon the Saddle–node system, estimated prediction accuracy of the time series envelope method and the DL model, respectively. (C) and (D): the Bautin system. (E) and (F): the Compost-Bomb system.}
\end{figure}

\clearpage
\begin{figure}
\centering
\includegraphics[width=0.7\textwidth]{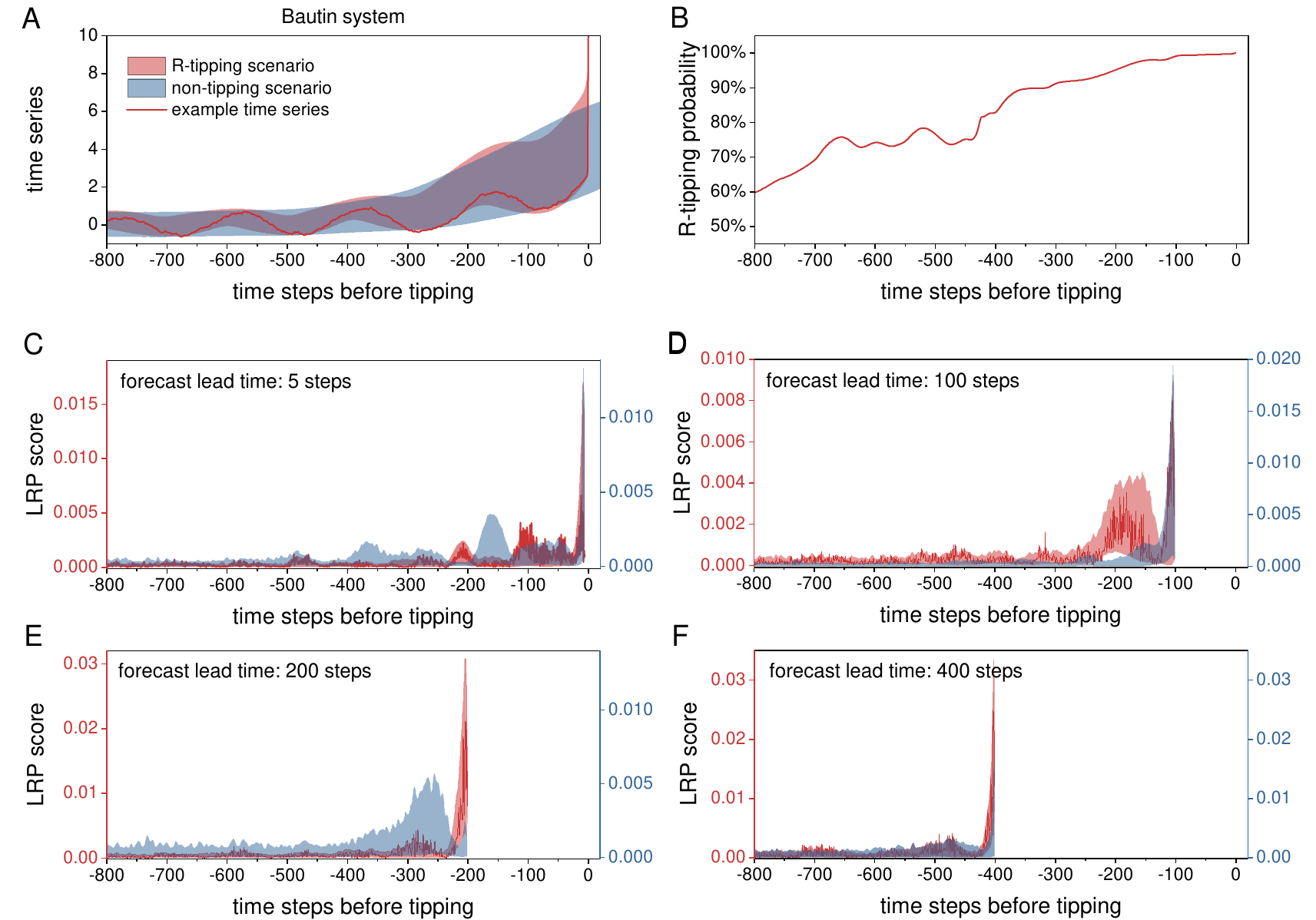}
\caption{The same configuration as Fig. 4, but for Bautin system.}
\end{figure}

\clearpage
\begin{figure}
\centering
\includegraphics[width=0.7\textwidth]{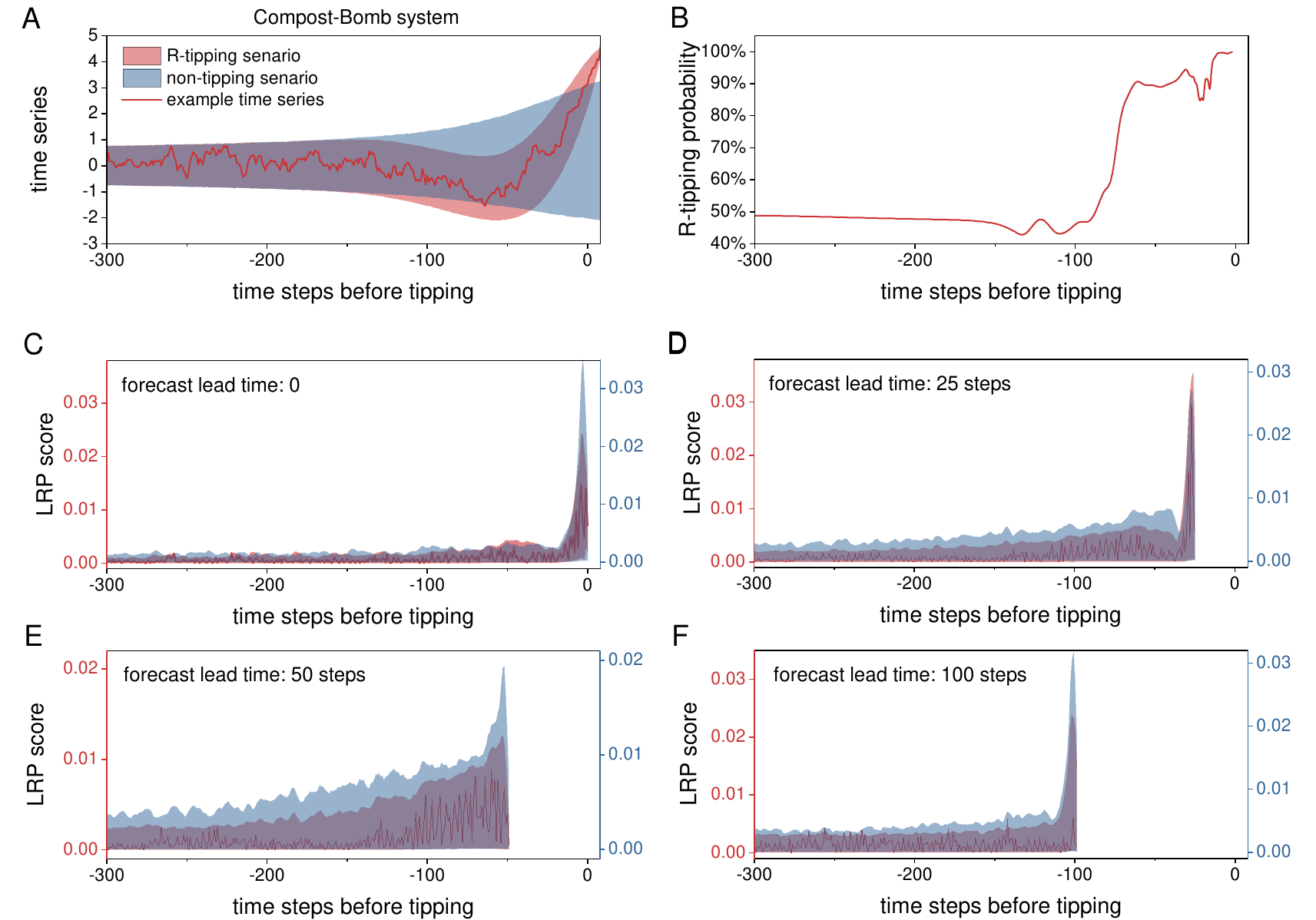}
\caption{The same configuration as Fig. 4, but for Compost-Bomb system.}
\end{figure}

\clearpage
\begin{figure}
\centering
\includegraphics[width=0.3\textwidth]{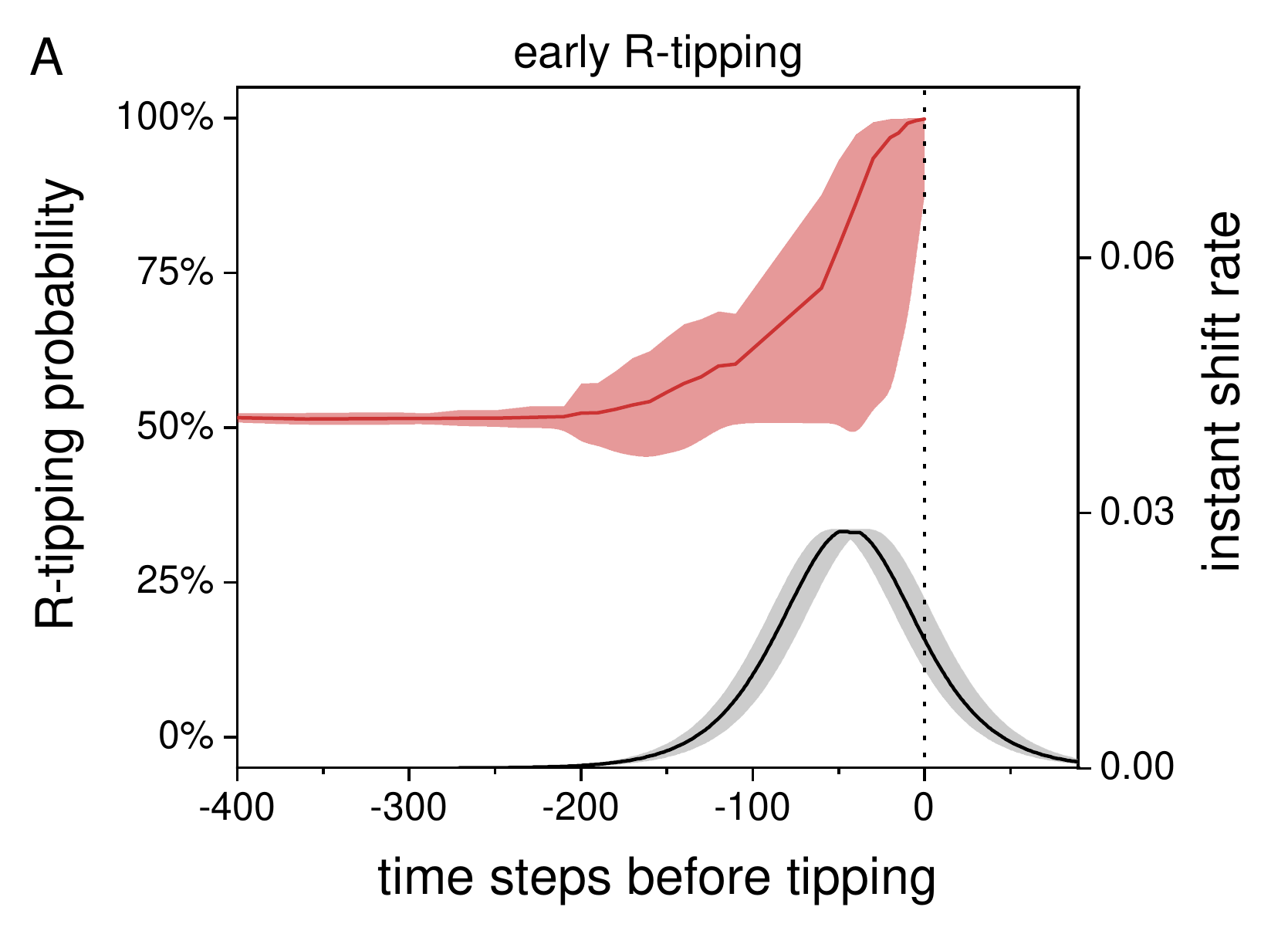}
\centering
\includegraphics[width=0.3\textwidth]{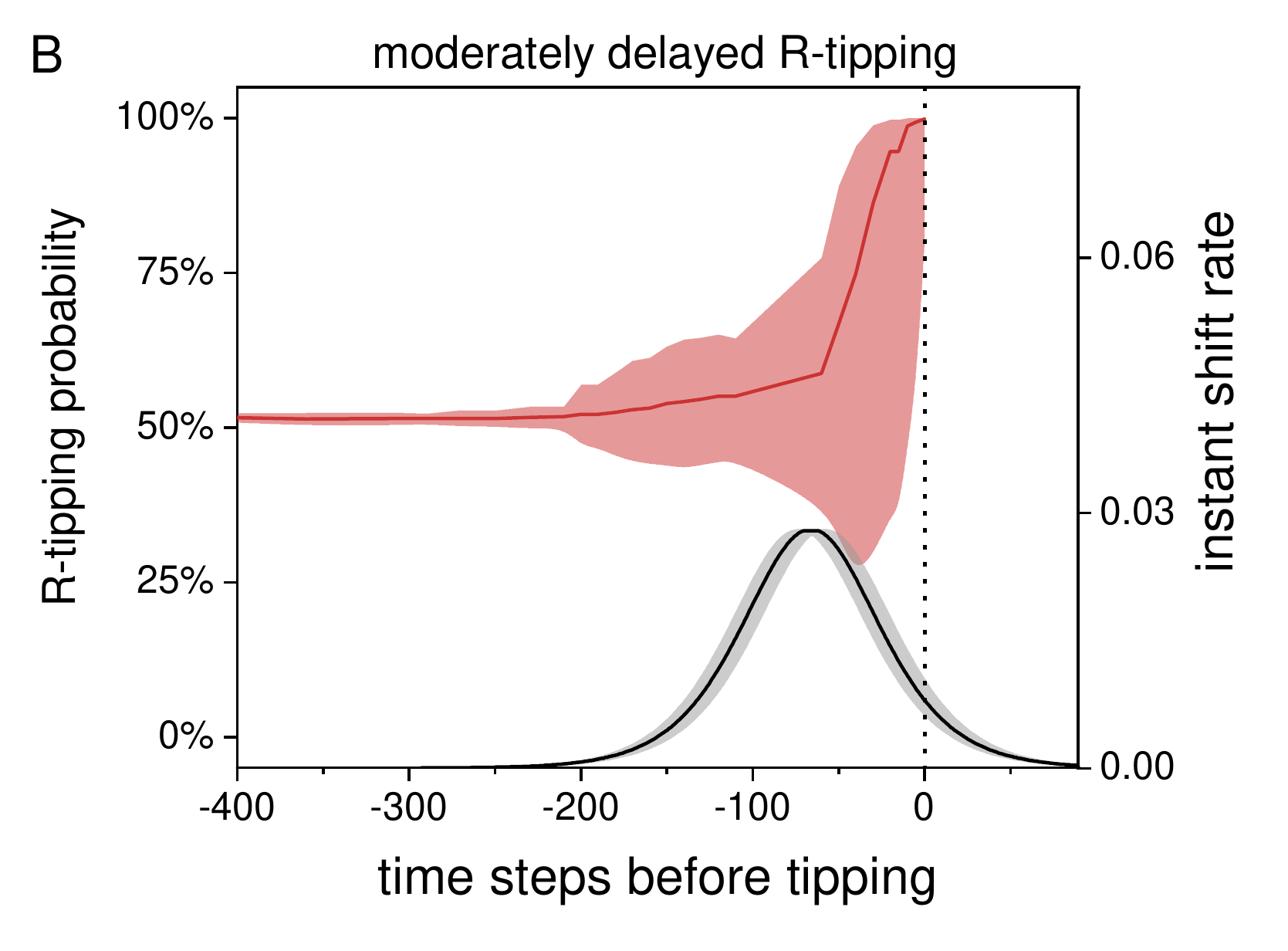}
\centering
\includegraphics[width=0.3\textwidth]{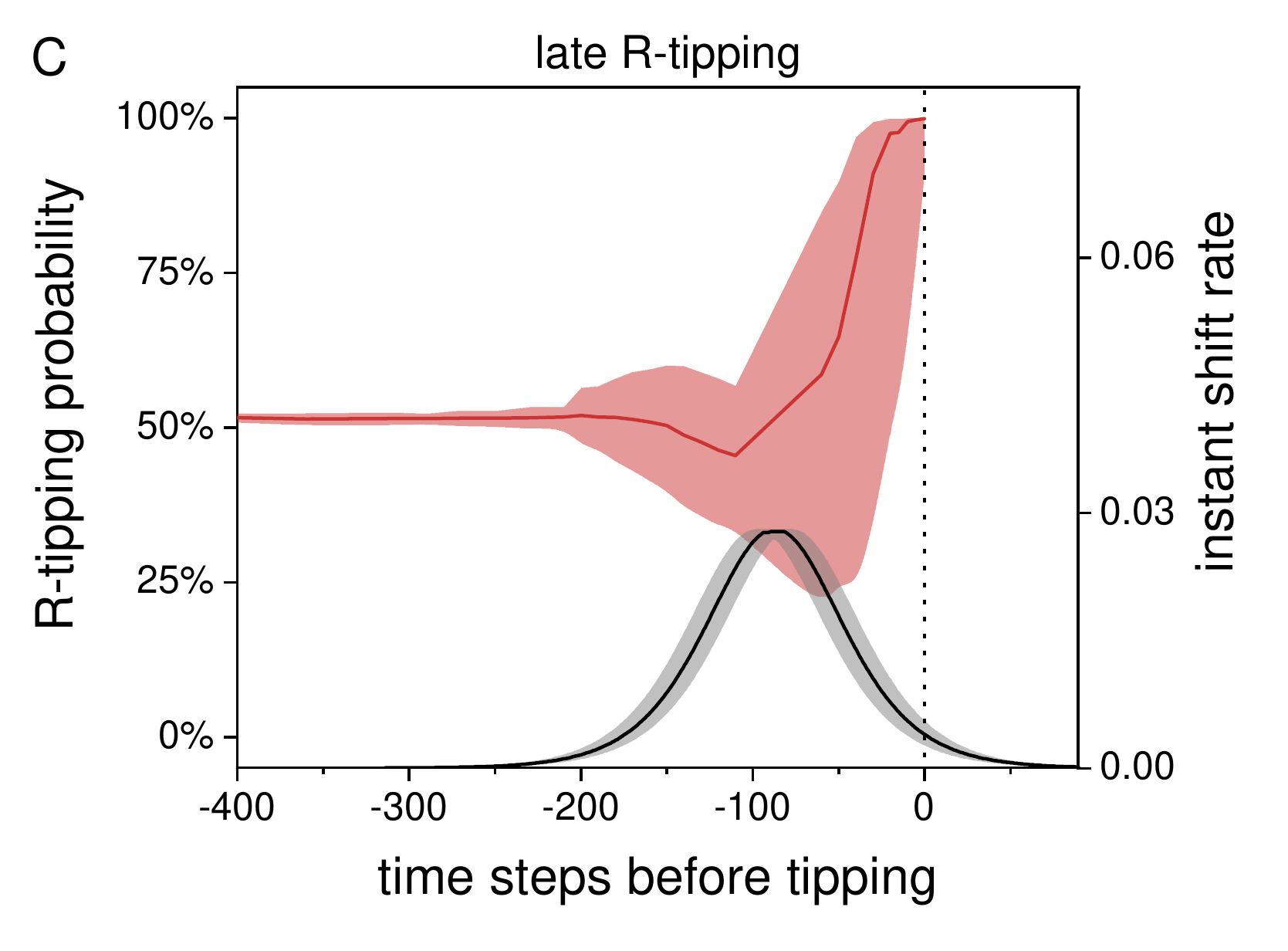}
\caption{The influence of early or late R-tipping on the DL-derived R-tipping probabilities of the Saddle–node system. The time series samples in group A are categorized into three equal quantiles based on their time delays relative to the maximum of the instant shift rate, and the DL-derived R-tipping probabilities within the quantiles are composited separately. Composite results for the first (A), the second (B) and the third quantiles (C) are shown. Red solid lines represents the mean values of the DL-derived R-tipping probabilities as a function of the time, and red shading areas for the 99\% confidence intervals. Black solid lines and grey shading areas are for the the instant shift rate of the forcing. }
\end{figure}

\clearpage
\begin{figure}
\centering
\includegraphics[width=0.3\textwidth]{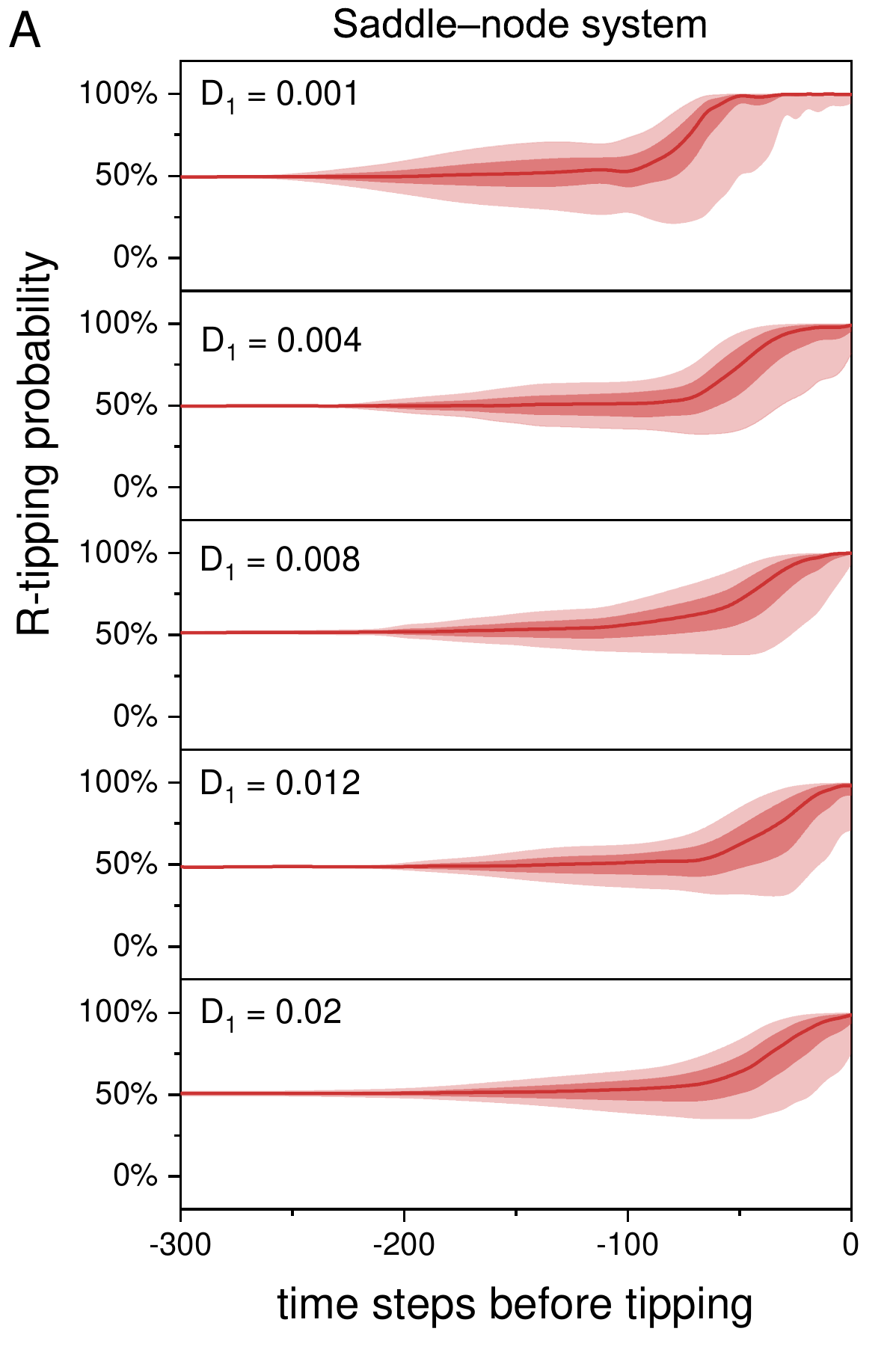}
\centering
\includegraphics[width=0.3\textwidth]{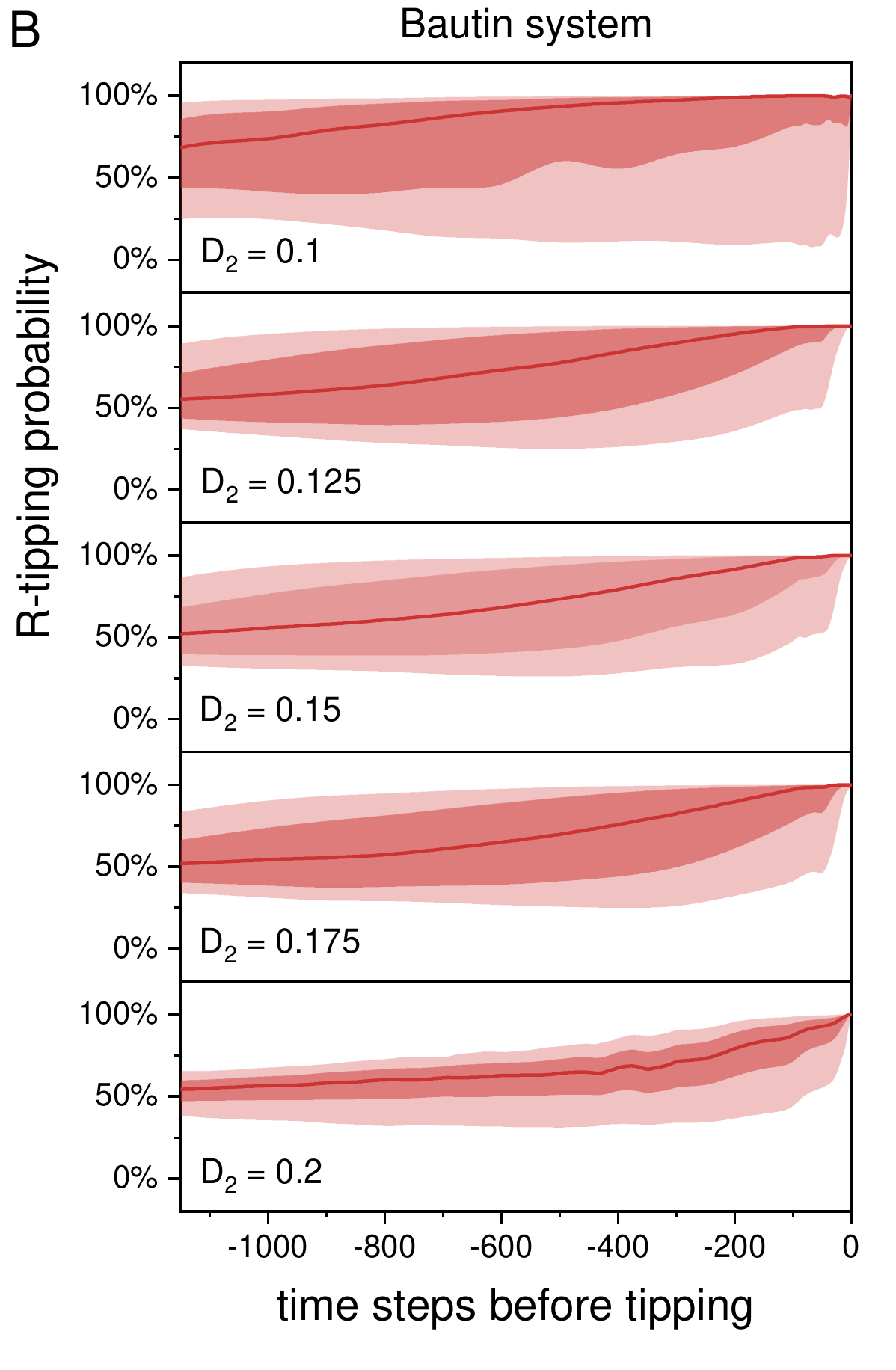}
\centering
\includegraphics[width=0.3\textwidth]{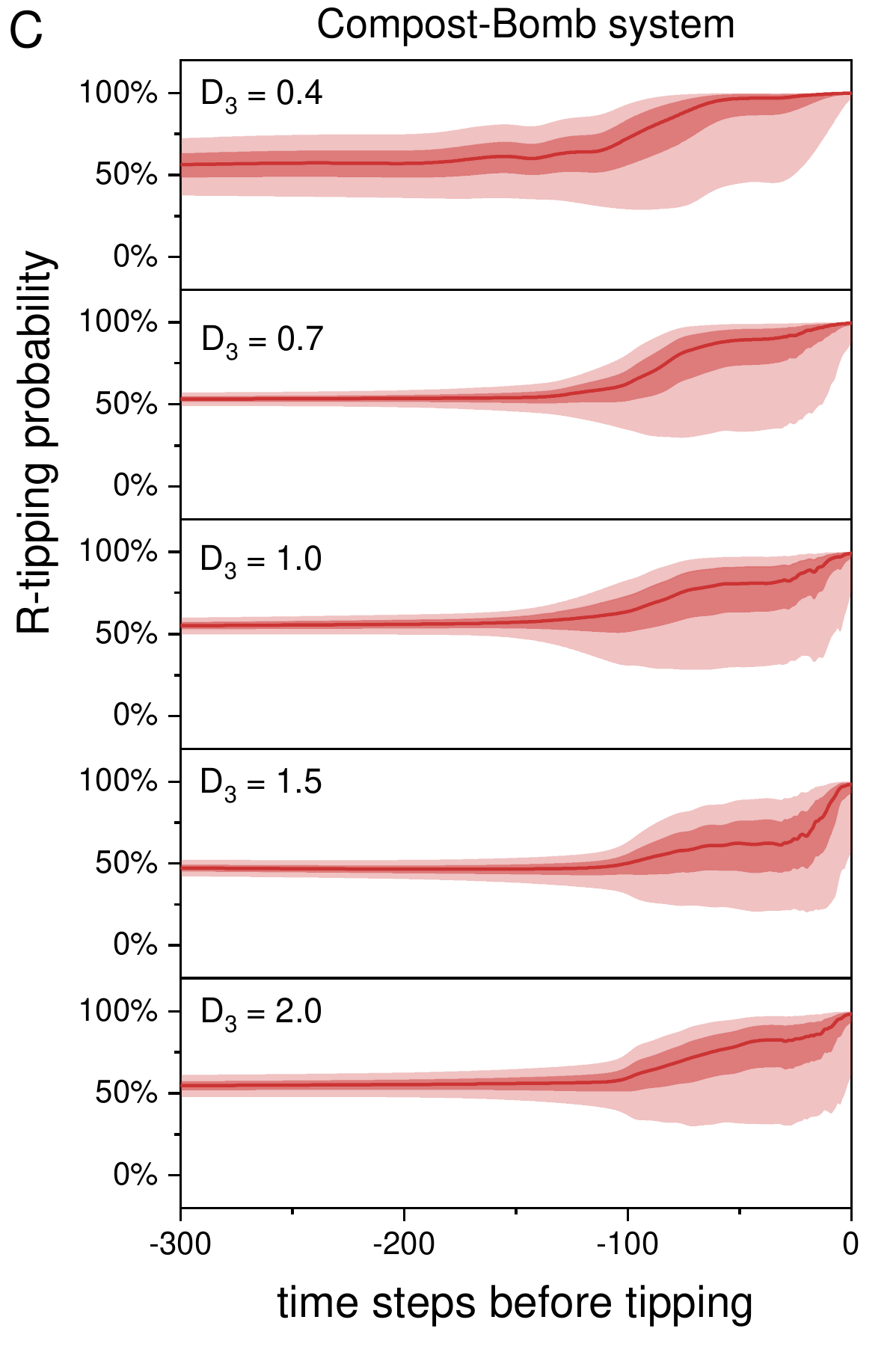}
\caption{The influence of noise perturbations' magnitudes on the DL-derived R-tipping probabilities. (A) Upon Saddle–node system: The red solid line represents the mean values of the DL-derived R-tipping probabilities for the time series manifesting R-tipping scenario, while the light red and red shading areas depict the 99\% and 75\% confidence intervals, respectively. $D_{1}$ denotes the magnitude of the noise perturbations in Saddle–node system (see Materials and methods). Panels from top to bottom denote the DL-derived R-tipping probabilities when $D_{1}$ is set as 0.001, 0.004, 0.008, 0.012, and 0.02, respectively. (B) mirrors the configuration of (A), but upon Bautin system, and $D_{2}$ is set as 0.1, 0.125, 0.15, 0.175, and 0.2, respectively. (C) mirrors the configuration of (A), but upon Compost-Bomb system, and $D_{3}$ is set as 0.4, 0.7, 1.0, 1.5, and 2.0, respectively.}
\end{figure}

\clearpage
\begin{figure}
\centering
\includegraphics[width=0.3\textwidth]{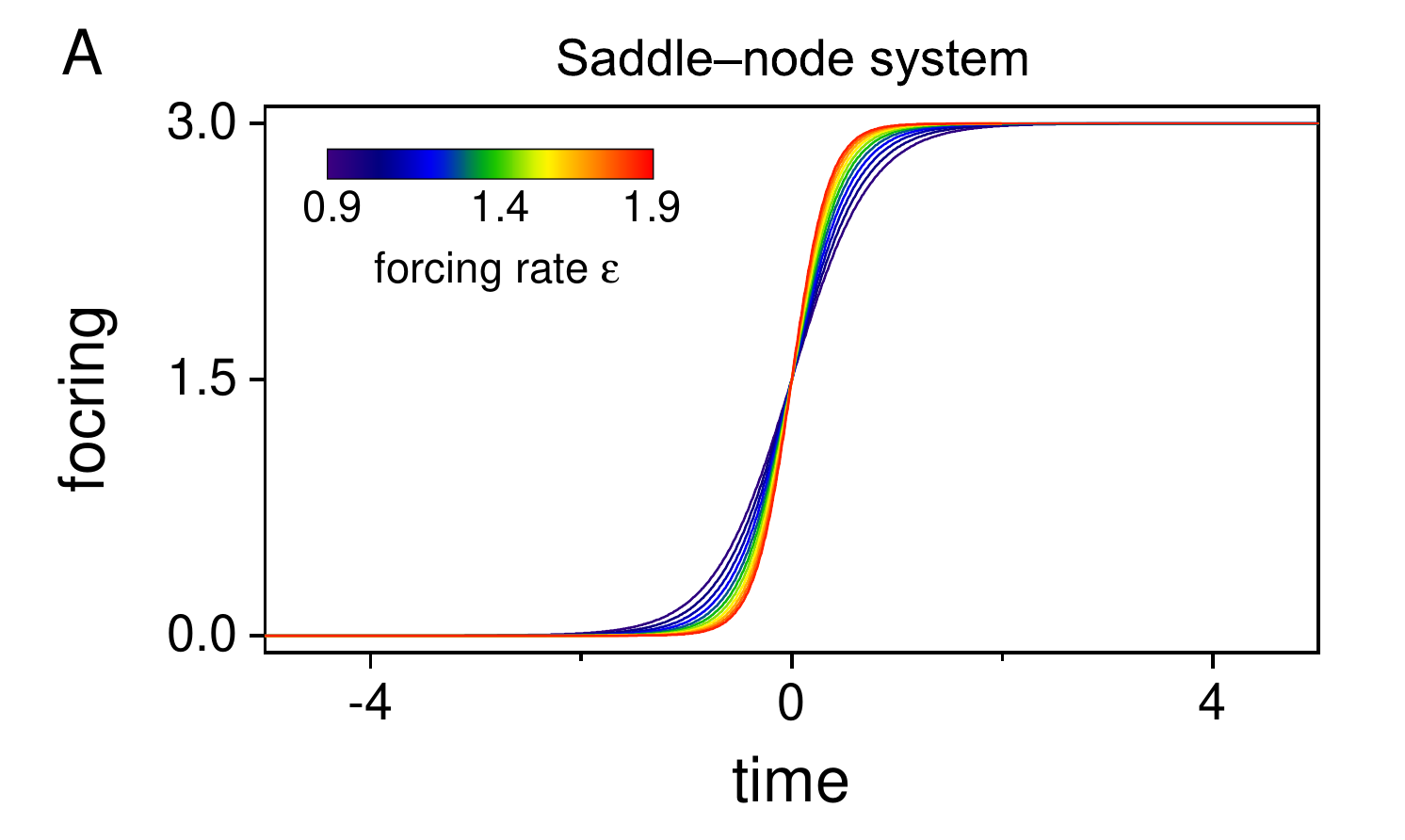}
\centering
\includegraphics[width=0.3\textwidth]{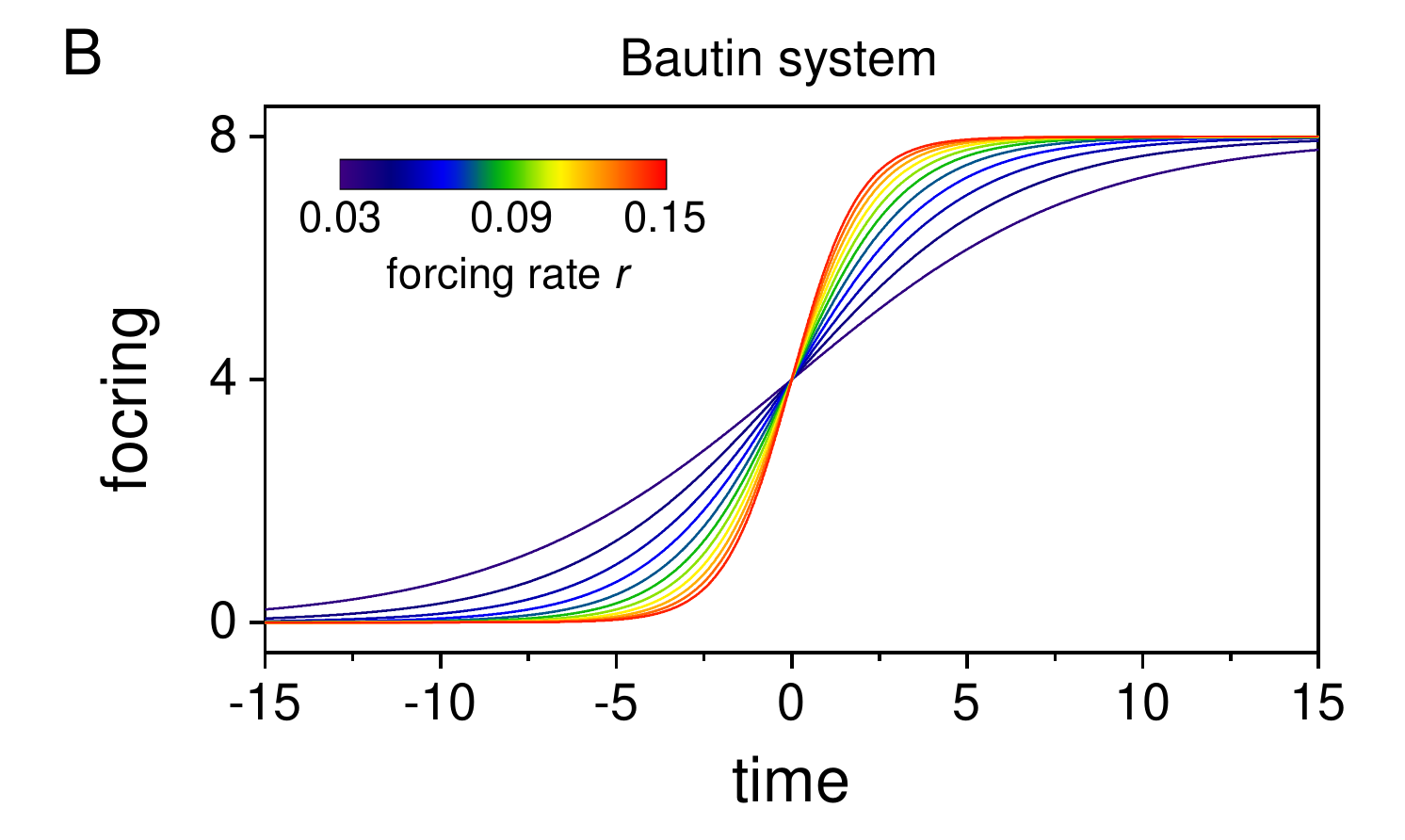}
\centering
\includegraphics[width=0.3\textwidth]{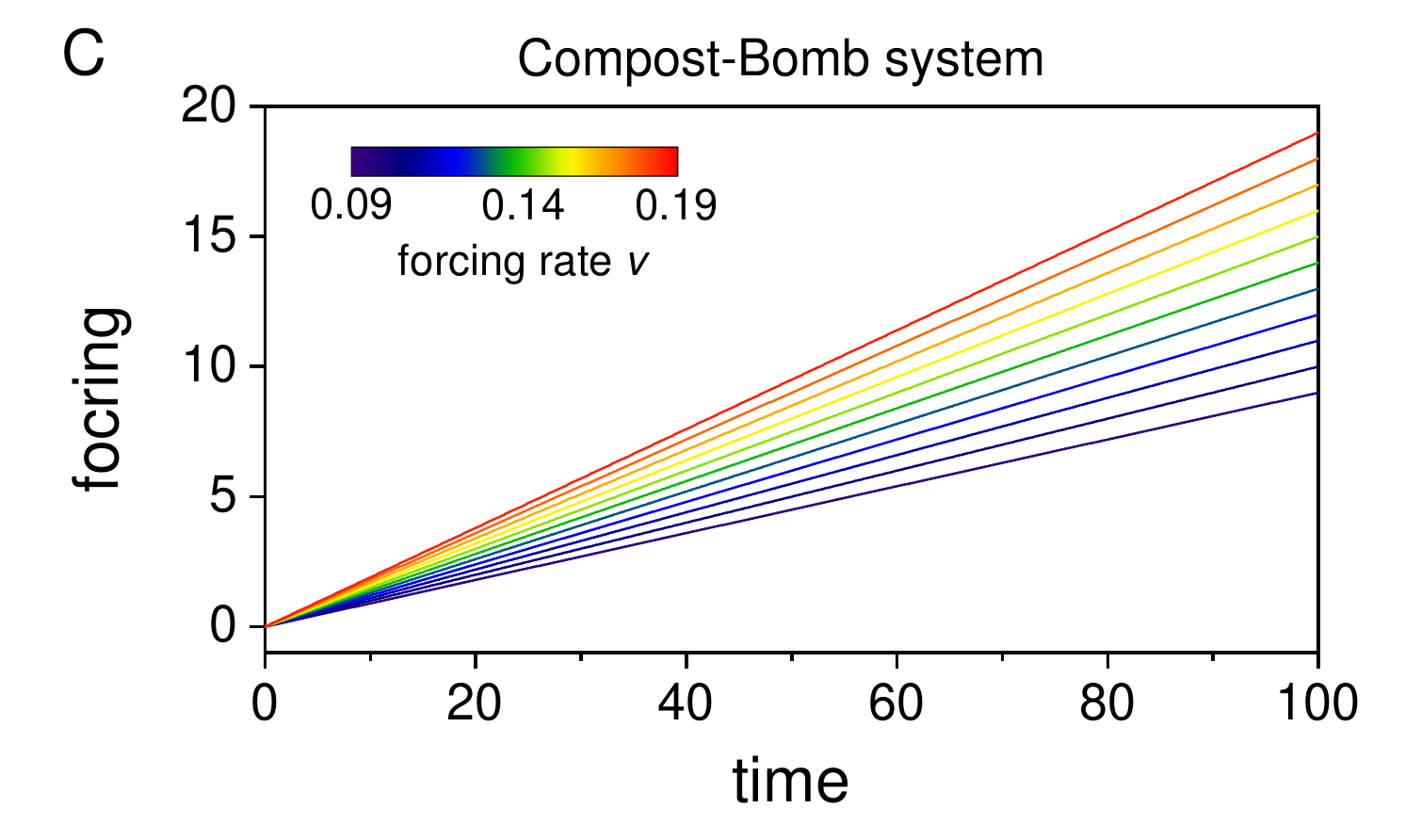}
\caption{Illustrating different forcing rates: time series of forcing in paradigmatic systems such as the Saddle-node system (A), Bautin system (B), and Compost-Bomb system (C). }
\end{figure}

\clearpage
\begin{table}\centering
\caption{Definitions and set values of the parameters for simulating Compost-Bomb system}

\begin{tabular}{lll}
Parameters & Definitions & Set values (unit) \\
\midrule
$\Pi$ & litter fall from plants & 1.055 ($kgm^{-2}yr^{-1}$) \\
$r_0$ & microbial respiration parameter & 0.01 ($yr^{-1}$) \\
$alpha$ & microbial respiration parameter & ln(2.5)/10 \\
$\mu$ & soil heat capacity & $2.5\times10^{6} (Jm^{-2}degC^{-1})$ \\
$A$ & specific heat from microbial respiration & $3.9\times10^{7} (Jkg^{-1})$ \\
$\lambda$ & soil-to-air heat transfer coefficient & $5.049\times10^{6} (Jyr^{-1}m^{-2}degC^{-1})$ \\
\bottomrule
\end{tabular}
\end{table}